\documentclass[sigconf]{acmart}
\AtBeginDocument{%
  }

\usepackage{multirow}
\usepackage{makecell}
\newcommand{\st}{\emph{s.t. }}
\newcommand{\eg}{\emph{e.g., }}
\newcommand{\ie}{\emph{i.e., }}

\newcommand{\eat}[1]{}
\ifodd 1

\newcommand{\TODO}[1]{{\color{red}TODO:{#1}}}
\newcommand\beftext[1]{{\color[rgb]{0.5,0.5,0.5}{BEFORE:#1}}}
\else

\newcommand{\TODO}[1]{}
\newcommand{\beftext}[1]{}
\fi

\copyrightyear{2025}
\acmYear{2025}
\setcopyright{acmlicensed}\acmConference[KDD '25]{Proceedings of the 31st ACM SIGKDD Conference on Knowledge Discovery and Data Mining V.1}{August 3--7, 2025}{Toronto, ON, Canada}
\acmBooktitle{Proceedings of the 31st ACM SIGKDD Conference on Knowledge Discovery and Data Mining V.1 (KDD '25), August 3--7, 2025, Toronto, ON, Canada}
\acmDOI{10.1145/3690624.3709323}
\acmISBN{979-8-4007-1245-6/25/08}

\begin{document}

\title{AutoSTF: Decoupled Neural Architecture Search for Cost-Effective Automated Spatio-Temporal Forecasting}

\author{Tengfei Lyu}
\affiliation{%
  \institution{The Hong Kong University of Science and Technology (Guangzhou)}
  \city{Guangzhou}
  \country{China}}
\email{tlyu077@connect.hkust-gz.edu.cn}

\author{Weijia Zhang}
\affiliation{%
  \institution{The Hong Kong University of Science and Technology (Guangzhou)}
  \city{Guangzhou}
  \country{China}}
\email{wzhang411@connect.hkust-gz.edu.cn}

\author{Jinliang Deng}
\affiliation{%
  \institution{The Hong Kong University of Science and Technology}
  \city{Hong Kong}
  \country{China}}
\email{dengjinliang@ust.hk}

\author{Hao Liu}
\authornote{Corresponding author.}
\affiliation{%
  \institution{The Hong Kong University of Science and Technology (Guangzhou)}
  \institution{The Hong Kong University of Science and Technology}
  \city{Guangzhou \& Hong Kong}
  \country{China}}
\email{liuh@ust.hk}


\begin{abstract}
Spatio-temporal forecasting is a critical component of various smart city applications, such as transportation optimization, energy management, and socio-economic analysis. Recently, several automated spatio-temporal forecasting methods have been proposed to automatically search the optimal neural network architecture for capturing complex spatio-temporal dependencies. However, the existing automated approaches suffer from expensive neural architecture search overhead, which hinders their practical use and the further exploration of diverse spatio-temporal operators in a finer granularity. In this paper, we propose AutoSTF, a decoupled automatic neural architecture search framework for cost-effective automated spatio-temporal forecasting. From the efficiency perspective, we first decouple the mixed search space into temporal space and spatial space and respectively devise representation compression and parameter-sharing schemes to mitigate the parameter explosion. The decoupled spatio-temporal search not only expedites the model optimization process but also leaves new room for more effective spatio-temporal dependency modeling. From the effectiveness perspective, we propose a multi-patch transfer module to jointly capture multi-granularity temporal dependencies and extend the spatial search space to enable finer-grained layer-wise spatial dependency search. Extensive experiments on eight datasets demonstrate the superiority of AutoSTF in terms of both accuracy and efficiency. Specifically, our proposed method achieves up to 13.48$\times$ speed-up compared to state-of-the-art automatic spatio-temporal forecasting methods while maintaining the best forecasting accuracy. The source code and data are available at \url{https://github.com/usail-hkust/AutoSTF}.
\end{abstract}



\maketitle

\section{Introduction}

Spatio-temporal forecasting is the process of predicting future states that depend on both spatial and temporal contexts~\cite{wang2020deep, jiang2023spatio, lin2022conditional}.
For example, in transportation systems, the future traffic flow and speed can be estimated by simultaneously learning the spatio-temporal dependencies of historical traffic conditions generated by geo-distributed roadside sensors~\cite{zhang2022crossformer,liu2020multi,fang2021mdtp}.
Accurate spatio-temporal forecasting plays a pivotal role in various smart city applications, such as human mobility modeling~\cite{li2022spatial, cirstea2022towards, HanDSFL021}, demand-supply rebalancing~\cite{zhang2020semi, ling2023sthan}, and urban anomaly event detection~\cite{tedjopurnomo2020survey,fliupractical2022}.

\begin{figure}[t]
  \centering
  \vspace{-0.cm}
  \includegraphics[width=1\linewidth]{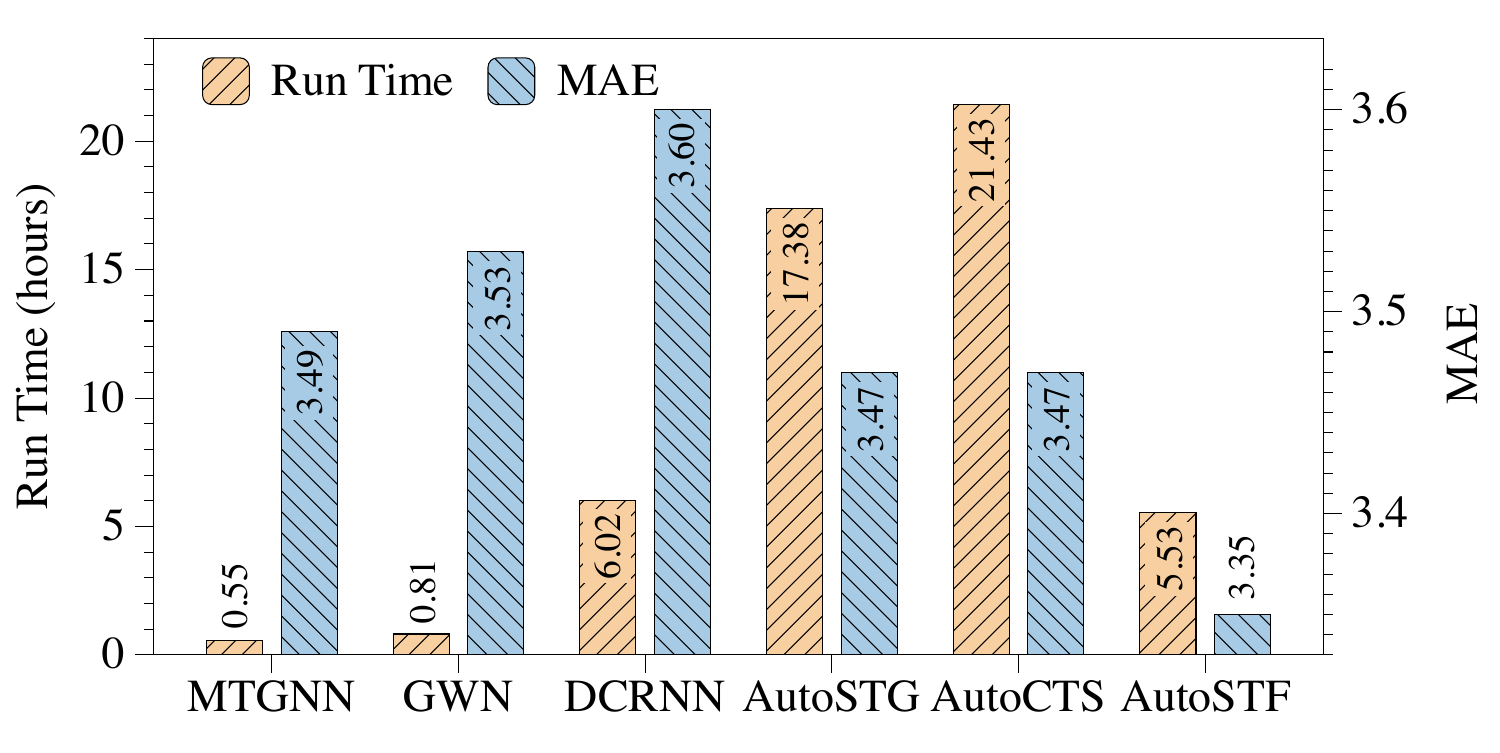}
  \vspace{-0.6cm}
  \caption{The training time and forecasting accuracy (mean squared error) comparison of manually-designed and automated spatio-temporal forecasting models on the METR-LA dataset. Our proposed method (AutoSTF) achieves the best forecasting accuracy while taking much less training time compared with all existing automated models.}
  \label{fig:Introduction_run_time}
  \Description{}
  \vspace{-0.4cm}
\end{figure}

In the past decade, extensive efforts have been made to capture spatio-temporal dependencies by leveraging advanced deep learning techniques~\cite{liang2022mixed, qu2022forecasting, fang2023spatio}.
To name a few, D$^2$STGNN~\cite{shao2022decoupled} incorporates Graph Neural Network (GNN) and Recurrent Neural Network (RNN) to model the diffusion process of traffic dynamics, while STEP~\cite{shao2022pre} devise pre-training to encode temporal patterns into segment-level representations. 
As another example, METRO~\cite{cui2021metro} is a versatile framework that utilizes a multi-scale temporal graph neural network to model dynamic and cross-scale variable correlations simultaneously.
Despite adopting various neural network blocks that have proven effective in preserving spatio-temporal dependencies, it requires substantial domain knowledge and extensive expert efforts to design an optimal model architecture~\cite{pan2021autostg,wu2021autocts}.

As an emerging trend, recently a few automated spatio-temporal methods have been proposed to determine the optimal neural architecture~\cite{pan2021autostg, wu2021autocts, wu2023autocts+}.
In general, such methods automatically search the model architecture by exploiting a pre-defined spatial and temporal operator space in an end-to-end manner.
However, the combinatorial nature of diverse spatio-temporal operators leads to a vast number of potential model architectures, making the search process computationally expensive and time-consuming.
As depicted in Figure~\ref{fig:Introduction_run_time}, existing automated spatio-temporal forecasting methods take an order of magnitude longer than manually-designed methods while yielding marginal prediction accuracy improvements.
The extensive computational overhead of automated approaches not only prevents their practical use in real-world tasks, but also hinders further optimization of the forecasting accuracy.
It is imperative to develop an efficient and effective automated spatio-temporal forecasting method. In fact, after analyzing the performance of automated methods on various spatio-temporal forecasting datasets, we identify two aspects that were previously overlooked by existing approaches, which provide great potential for us to improve the efficiency and effectiveness of automated spatio-temporal forecasting. We detail our research insights below.

\emph{Mixed spatio-temporal search space.} Existing automated models search for the optimal neural architecture within a unified search space, which combines various temporal and spatial operators. From an efficiency perspective, such a mixed space entangles the search process for spatial and temporal correlation modeling blocks, resulting in exponential search complexity. However, current spatio-temporal forecasting models typically capture spatio-temporal correlations separately, \eg using GNNs to capture spatial dependency and RNNs for temporal dependency~\cite{lai2023lightcts}. Searching for a forecasting model in a mixed search space may lead to high computational redundancy and potentially suboptimal neural architecture due to insufficient exploration.
Thus, we propose a decoupled spatio-temporal search framework wherein the spatial dependency block and temporal dependency block are determined separately. Moreover, we devise a representation compression scheme to distill key temporal knowledge and introduce a parameter-sharing scheme to reduce the optimization overhead of the searched spatial dependency network.
By decoupling the mixed search space and minimizing the parameter size, the search process can be significantly accelerated, allowing for more flexible exploration in both the parameter optimization and advanced search space design.

\emph{Coarse-grained spatio-temporal correlation search.} Limited by computational inefficiency, current automated forecasting methods typically leverage spatio-temporal dependencies at a coarse granularity. 
In the temporal domain, previous works have commonly applied the same spatial operator to all previous $T$ time steps, disregarding the possibility that the correlation may vary across different time steps. 
Furthermore, existing studies have adopted an identical message-passing scheme in different GNN layers, which overlooks distinct spatial dependencies in different multi-hop neighbors.
We detail the effectiveness bottleneck with more empirical evidence in Section~\ref{sec:The Effectiveness Bottleneck}.
In this paper, we propose a multi-patch transfer module that divides temporal representations into different patches so that time-varying pair-wise dependency can be preserved in different patches.
Besides, we introduce the spatial adjacency matrix as a new class of spatial operators and search the adjacency for each layer to strengthen the spatial dependency modeling.
By incorporating fine-grained temporal and spatial dependencies, the accuracy of spatio-temporal forecasting can be further improved.

Along these lines, in this paper, we develop \textbf{AutoSTF}, a cost-effective decoupled \textbf{Auto}mated \textbf{S}patio-\textbf{T}emporal \textbf{F}orecasting framework.
Our main contributions are summarized as follows.
From the efficiency perspective, we propose a decoupled automated spatio-temporal forecasting framework to reduce the neural architecture search overhead. By incorporating representation compression and parameter sharing schemes, AutoSTF not only expedites the model optimization process but also leaves new room for more effective spatio-temporal forecasting.
From the effectiveness perspective, we propose a multi-patch transfer module and layer-wise message-passing spatial operators to respectively capture fine-grained temporal and spatial dependencies, thereby enhancing the forecasting accuracy.
We conduct extensive experiments on eight datasets from different application domains to demonstrate that our proposed framework outperforms state-of-the-art automated spatio-temporal forecasting models in terms of both efficiency and effectiveness.


\begin{figure*}[t]
  \centering
  \includegraphics[width=\linewidth]{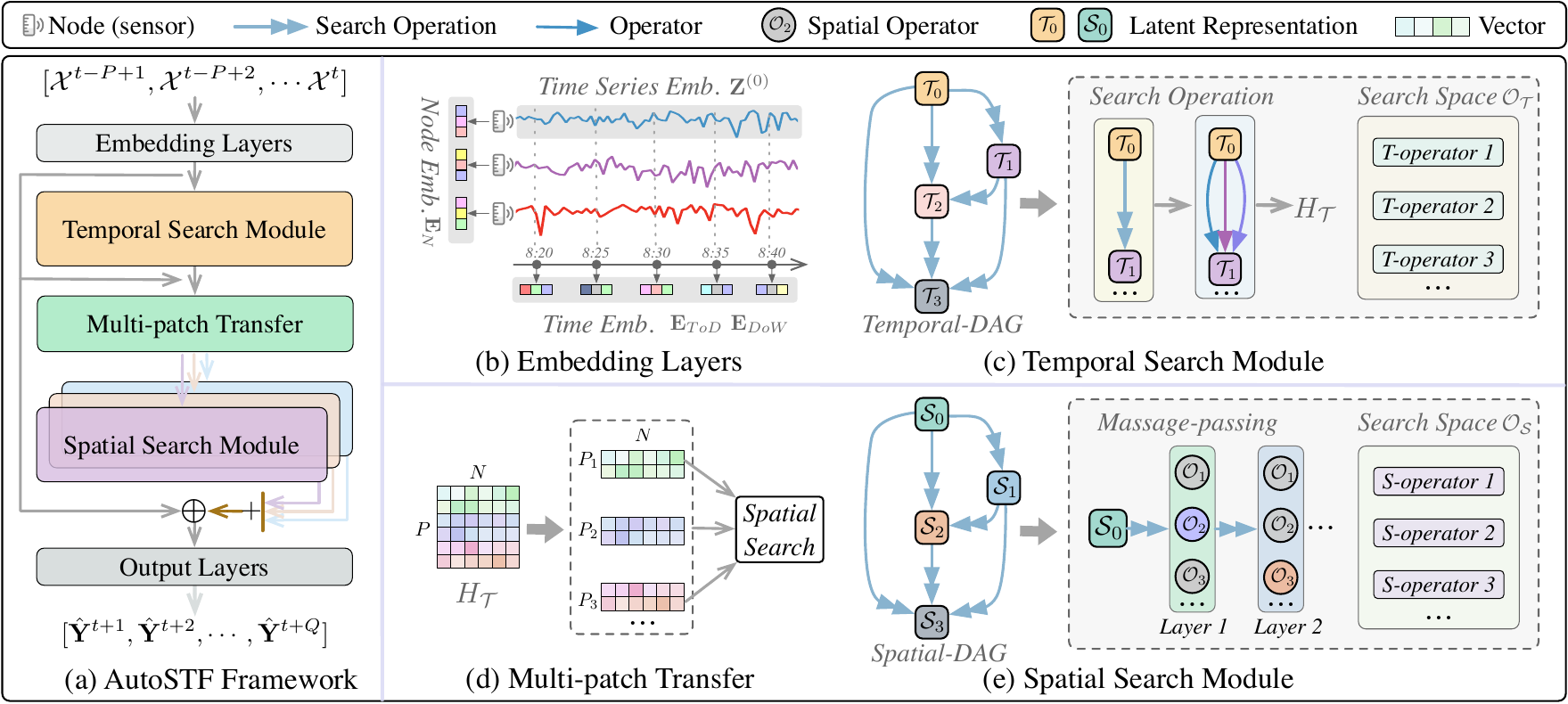}
  \vspace{-0.4cm}
  \caption{AutoSTF framework.
  (a) shows the overview framework of AutoSTF. (b) depicts the Embedding Layers, which consist of raw time series embedding, node embedding, and time embedding. (c) illustrates the Temporal Search Module, which searches for the optimal temporal operator within the Temporal-DAG to model complex temporal dependencies. (d) describes the Multi-patch Transfer, which segments the embedding into several patches along the temporal feature axis and compresses each into a dense semantic representation. (e) presents the Spatial Search Module, tasked with searching for the optimal spatial operator and integrating it with the preceding temporal dependencies to uncover fine-grained spatial-temporal correlations.
  The two-headed arrow denotes the search operation. The single-headed arrow denotes the operator, and different colors represent different operators.}
  \label{fig:framework}
  \Description{AutoSTF Framework.}
  \vspace{-0cm}
\end{figure*}

\section{Preliminaries}
\label{sec:Preliminaries}

\begin{definition}[\textbf{Spatial Graph}]
We denote the spatial graph as $\mathcal{G} = (\mathcal{V}, \mathcal{E})$, where $|\mathcal{V}|=N$ is the set of nodes (sensors), indicating each node in $\mathcal{V}$ corresponds to a time series; $N$ is the number of nodes; and $\mathcal{E}$ denotes the set of edges that represent spatial correlations between different nodes. $\mathbb{A} \in \mathbb{R}^{N \times N}$ is the adjacency matrix of spatial graph $\mathcal{G}$.
\end{definition}

In this work, the adjacency matrix $\mathbb{A}$ can be predefined, \eg based on the node distances, or adaptive by a data-driven approach, \eg based on time series or node embedding. 
The time series on the spatial graph can be denoted as graph signal matrix.

\begin{definition}[\textbf{Graph Signal Matrix}]
We use $\mathcal{X} \in \mathbb{R}^{T \times N \times C}$ to denote the graph signal matrix, where $T$ denotes the number of time steps, and $C$ denotes the number of features, \eg traffic speed. 
\end{definition}

\textbf{Spatio-temporal Forecasting.}
We consider both single-step and multi-step spatio-temporal forecasting.
Given $\mathcal{G}$, $\mathcal{X}$, and the historical time steps $P$, the goal of multi-step forecasting is to predict the value at all $Q$ future time steps:
\begin{equation}
(\hat{\mathbf{Y}}^{t+1}, \hat{\mathbf{Y}}^{t+2}, \cdots ,\hat{\mathbf{Y}}^{t+Q})  \leftarrow \mathcal{F}(\mathbf{H}^{t-P+1}, \mathbf{H}^{t-P+2}, \cdots \mathbf{H}^{t}),
\end{equation}
where $\mathbf{H}^{t} = (\mathcal{G}^{t}, \mathcal{X}^{t})$, and $\hat{\mathbf{Y}}^{t}$ denotes the predictive value at time step $t$; $\mathcal{F}$ denotes the forecasting model. 
While single-step forecasting aims to predict the value of $Q$-th future time step.

\textbf{Problem Statement.}
In this work, we aim to design an efficient and effective automated spatio-temporal forecasting model $\mathcal{F}$. 
We try to search for an optimal neural architecture $M^{*}$ that contains both architecture parameters $\Theta$ and model parameters $\omega$. 
The objective function is as follows:
\begin{equation}
\begin{aligned}
M^{*} = \underset{\Theta}{\min} \mathcal{L}(\omega^{*}, \Theta, \mathbb{D}_{val}), \\
\st \omega^{*} = \arg \underset{\omega}{\min} \mathcal{L}(\omega, \Theta, \mathbb{D}_{train}),
\end{aligned}
\end{equation}
where $M^{*} \in \mathcal{M}$ denotes the optimal neural architecture after the search phase, and $\mathcal{M}$ denotes the search space of model architecture. $\mathcal{L}$ is the loss function. 
$\mathbb{D}_{train}$ and $\mathbb{D}_{val}$ are the training dataset and validation dataset, respectively.

\begin{table}[b] \footnotesize
\setlength{\tabcolsep}{2.5pt} 
  \caption{The temporal and spatial operators of search space.}
  \vspace{-0.2cm}
  \label{tab:all_operators}
  \begin{tabular}{c|l|l}
    \toprule
    Type & Operators & Formulary \\
    
    \midrule
    \multirow{2}*{$\mathcal{T}$} & $GDCC$ & $GDCC(X) = XW_1\odot \sigma(XW_2))$ \\
    & $Informer$ & $Informer(X) = Softmax(\frac{(XW_{Q})(XW_{K})^{T}}{\sqrt{D^{\prime}}}(XW_V))$ \\

    \midrule
    \multirow{4}*{$\mathcal{S}$} & $GNN_{fixed}$ & $GNN_{fixed}(X) = \sum_{k=0}^{K} \mathbf{A}_{dis}^{k}XW_{fixed}$ \\
    & $GNN_{adap}$ & $GNN_{adap}(X) = \sum_{k=0}^{K}\mathbf{A}_{adap}^{k}XW_{adap}$ \\
    & $GNN_{att}$ & \makecell[r]{$GNN_{att}(X) = (ReLU(\mathbf{A}_{att}XW_i^V\cdot W_1+b_1))\cdot W_2+b_2$ \\ 
    $s.t. \mathbf{A}_{att} = softmax(\frac{XW_t^Q\cdot(XW_i^K)^T}{\sqrt{D^{\prime}}})$} \\
    \bottomrule
  \end{tabular}
\end{table}

\section{The AutoSTF Framework}
\label{sec:Automated Spatio-temporal Forecasting}

In this section, we provide a comprehensive overview of our proposed model. Additionally, we give a detailed analysis in Appendix \ref{Appendix:sec:Model Design and Analysis} to explain why the model is designed in this way. We list all the operators of temporal search and spatial search in Table \ref{tab:all_operators}.

\subsection{Embedding Layers}
\label{sec:embedding_layer}
Here we introduce the three kinds of embeddings preprocessed in the Embedding Layers, including time series embedding, node embedding, and time embedding.

The original time series $\mathcal{X} \in \mathbb{R}^{T\times N\times C}$ is processed through a linear layer to obtain an initial latent representation:
$\mathbf{Z}^{(0)} = \mathcal{X} \mathbf{W}_{t} + b_t$,
where $\mathbf{Z}^{(0)} \in \mathbb{R}^{T\times N\times D}$ is time series embedding, and $\mathbf{W}_{t}$ denotes the learnable matrix and $D$ is the hidden dimension.

Node embedding aims to identify and encode the spatial locations of different sensors and can be expressed as $\mathbf{E}_{N} \in \mathbb{R}^{N\times D}$.
Time embedding is designed to map the inherent time feature that can be extracted from raw time series.
$\mathbf{E}_{ToD} \in \mathbb{R}^{N_d \times D}$ and $\mathbf{E}_{DoW} \in \mathbb{R}^{N_w \times D}$ denote the embedding of day-of-week and time-of-day, respectively. $N_d$ denotes the time slots in a day, and $N_w$ denotes the day-of-week. 

We concatenate these three embeddings and feed it into the fully connected layers, as shown
$\mathbf{E}_{emb} = FC(concat[\mathbf{E}_{ToD}, \mathbf{E}_{DoW}, \mathbf{E}_N])$,
where $FC$ denotes the fully connected layers and $\mathbf{E}_{emb} \in \mathbb{R}^{N\times D}$ represents the corresponding embedding, which will be fully utilized in subsequent processes.

\subsection{Temporal Search Module}
\label{sec:temporal_search}

In this section, we first introduce the temporal search space and temporal-DAG, then explain how to address the temporal search.

\textbf{Temporal search space.}
The temporal search operator should prioritize two key factors: the capacity to accurately model both short-term and long-term temporal dependencies and the efficiency of temporal operators \cite{wu2021autocts}. 
Based on these findings, we select the Gated Dilated Causal Convolution (GDCC) \cite{diao2019dynamic,huang2020lsgcn} and informer~\cite{zhou2021informer} to construct the temporal search space. 
Furthermore, to enhance the flexibility of temporal search, we also incorporate two commonly used operators, namely $Zero$ and $Identity$, into our temporal search space, similar to the works in \cite{pan2021autostg,wu2021autocts,liu2018darts}. 
We define the temporal search space as $\mathcal{O}_{\mathcal{T}} = \{GDCC, Informer, Zero, Identity\}$.

\textbf{Temporal-DAG.}
In line with previous works \cite{liu2018darts,pan2021autostg,ke2023autostg+,wu2021autocts}, we have also employed a directed acyclic graph (DAG) to facilitate the search for various combinations of temporal operators, as shown in Figure \ref{fig:framework} (c).
The temporal-DAG has $N_{\mathcal{T}}$ nodes, and each node denotes a latent representation. 
$\mathcal{T}_0 = \mathbf{Z}^{(0)}$ represents the output from the Embedding Layer and serves as the input for this temporal-DAG. 
The edge in temporal-DAG denotes the search operation \cite{pan2021autostg} and will search for an optimal temporal operator in the temporal search space.
Specifically, for the edge between node $\mathcal{T}_0$ and $\mathcal{T}_1$, we have $|\mathcal{O}_{\mathcal{T}}|$ choice to determine a temporal operator.
The latent representation of $\mathcal{T}_i$ can only be transmitted to $\mathcal{T}_j$ using the selected operator, if $i<j$. 
In other words, the latent representation of the current node is enhanced by its previous nodes in the temporal-DAG.
For example, in Figure \ref{fig:framework} (c), the latent representation of node $\mathcal{T}_3$ is aggregated with the information from nodes $\mathcal{T}_0$, $\mathcal{T}_1$, and $\mathcal{T}_2$.

\textbf{Parameterizing temporal-DAG.}
Following DARTS framework \cite{liu2018darts}, we perform the temporal search within a temporal-DAG, as shown in Figure \ref{fig:framework}(c). 
Each operator in the temporal search space is capable of transforming a latent representation (\eg $\mathcal{T}_0$) into a new one (\eg $\mathcal{T}_1$) in the temporal-DAG.
We denote $\alpha$ as the architecture parameters of temporal-DAG and introduce the search operation between node $\mathcal{T}_i$ and $\mathcal{T}_j$ with ${\alpha}^{(i\rightarrow j)} \in \mathbb{R}^{|\mathcal{O}_{\mathcal{T}}|}$. 
Specifically, the weight of the operator $o \in \mathcal{O}_{\mathcal{T}}$ between node $\mathcal{T}_i$ and $\mathcal{T}_j$ is formulate as follow:
\begin{equation}
\begin{aligned}
{\alpha}^{(i\rightarrow j)}_{o} = \frac{exp(\alpha^{(i\rightarrow j)}_{o})}{\sum_{o^\prime \in \mathcal{O}_{\mathcal{T}}} exp(\alpha_{o^\prime}^{(i\rightarrow j)})},
\end{aligned}
\end{equation}
where $o$ represents an operator, \eg $Informer$, and $\alpha^{(i\rightarrow j)}_{o} \in \mathbb{R}^{}$ is a weight parameter of operator $o$. 

After calculating the weight of each operator between node $\mathcal{T}_i$ and $\mathcal{T}_j$, we can obtain the latent representation $\mathcal{T}^{(i\rightarrow j)}$ by computing the weighted sum of all operators between node $\mathcal{T}_i$ and $\mathcal{T}_j$ in the temporal-DAG. The formula is as follows:
\begin{equation}
\begin{aligned}
{\mathcal{T}}_j =\sum_{o \in \mathcal{O}_{\mathcal{T}}} {\alpha}^{(i\rightarrow j)}_{o} o(\mathcal{T}_i),
\end{aligned}
\end{equation}
where $o(\mathcal{T}_i)$ denotes the latent representation generated by applying operator $o$ with input $\mathcal{T}_i$.

Finally, we can obtain the latent representation $H_\mathcal{T}$ of the temporal-DAG. The formula is as follows:
\begin{equation}
\begin{aligned}
{H}_{\mathcal{T}} = \sum_{j=1}^{N_\mathcal{T}} \sum_{i=0}^{i<j} {\mathcal{T}}_j,
\end{aligned}
\end{equation}
where $N_\mathcal{T}$ is the number of node in the temporal-DAG, and ${H}_{\mathcal{T}} \in \mathbb{R}^{N\times P\times D}$ denotes the outputs embedding of the temporal search.
By performing the temporal search, we can automatically select the optimal operator to model temporal dependencies accordingly.

\begin{figure}[t]
  \centering
  \includegraphics[width=1\linewidth]{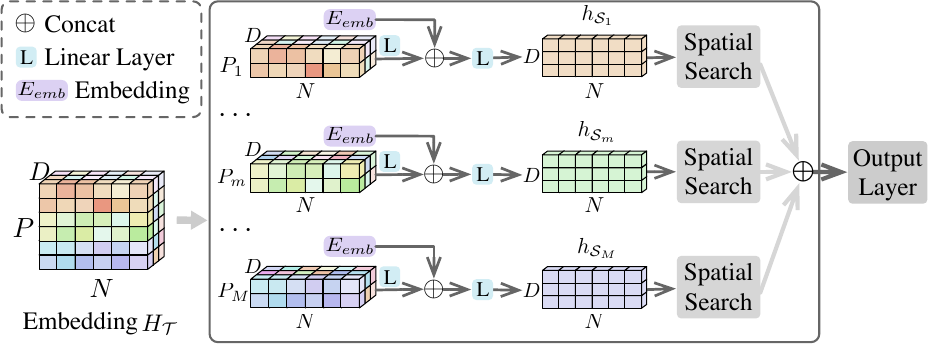}
  \vspace{-0.5cm}
  \caption{Multi-patch transfer module. The embedding $H_{\mathcal{T}}$ is the output of the Temporal Search Module. $P$ represents historical time steps segmented into $M$ patches.}
  \label{fig:multi-scale}
  \Description{Multi-patch transfer module}
  \vspace{-0.2cm}
\end{figure}

\subsection{Multi-patch Transfer Module}
\label{sec:multi-scale_transfer}
In this module, we segment the output embedding of the temporal search into several patches along the temporal feature axis and compress each patch into a dense semantic representation.
Implementing these strategies can explore the finer-grained spatial and temporal dependencies in subsequential spatial search, and decrease redundant computations in the temporal feature dimension.

Specifically, after obtaining the temporal search embedding ${H}_{\mathcal{T}} \in \mathbb{R}^{N\times P\times D}$ from the temporal search, we divide the temporal feature ( the $P$ dimension of ${H}_{\mathcal{T}}$) into several patches to explore the finer-grained spatio-temporal dependencies and effective compress the temporal feature for spatial search.
As illustrated in Figure \ref{fig:multi-scale}, we initially divide the temporal embedding ${H}_{\mathcal{T}}$ into $M$ patches:
\begin{equation}
\begin{aligned}
{H}_{\mathcal{T}} = \{h_{P_1}, h_{P_2}, \cdots h_{P_M} \},
\end{aligned}
\end{equation}
where $h_{P_m} \in \mathbb{R}^{N\times \frac{P}{M} \times D}$ denotes embedding of one patch and $M$ is the number of patches.
To reduce the redundant temporal information in subsequent spatial search, we compress the temporal feature with a linear layer, as shown in the following:
\begin{equation}
\begin{aligned}
{h}_{P_{m}}^{\prime} = h_{P_m}W_l + b_l,
\end{aligned}
\end{equation}
where $W_l \in \mathbb{R}^{\frac{P}{M} \times 1}$ is the learnable weights, and ${h}_{P_{m}}^{\prime} \in \mathbb{R}^{N\times 1\times D} = \mathbb{R}^{N\times D}$ denotes the compressed embedding of one patch.

To more precisely capture the finer-grained spatio-temporal dependencies, we concat embedding $E_{emb}$ with ${h}_{P_{m}}^{\prime}$, where embedding $E_{emb}$ comprises both time embedding and node embedding (refer to Section \ref{sec:embedding_layer}):
${h}_{\mathcal{S}_{m}} = Linear({h}_{P_{m}}^{\prime} \oplus E_{emb})$,
and ${h}_{\mathcal{S}_{m}} \in \mathbb{R}^{N\times D}$ is then used as the input embedding for the spatial search.
Finally, the output of multi-patch transfer can be denoted as ${H}_{\mathcal{P}} = \{ {h}_{\mathcal{S}_1}, {h}_{\mathcal{S}_2}, \cdots {h}_{\mathcal{S}_M} \}$.

The multi-patch transfer module serves as the bridge between temporal search and spatial search.
It involves segmenting the temporal feature and conducting the spatial search for each patch to investigate fine-grained spatio-temporal dependencies.
It converts the temporal search embedding $H_{\mathcal{T}}$ from $\mathbb{R}^{N\times P\times D}$ to $M \times \mathbb{R}^{N\times D}$, resulting in a significant reduction in computation time (refer to Table \ref{tab:decouple_space}). 
In addition, it combines the embedding $E_{emb}$ for the input of each spatial search, which preserves the temporal information and improves the accuracy of modeling spatio-temporal dependencies.

\begin{figure}[t]
  \centering
  \includegraphics[width=0.95\linewidth]{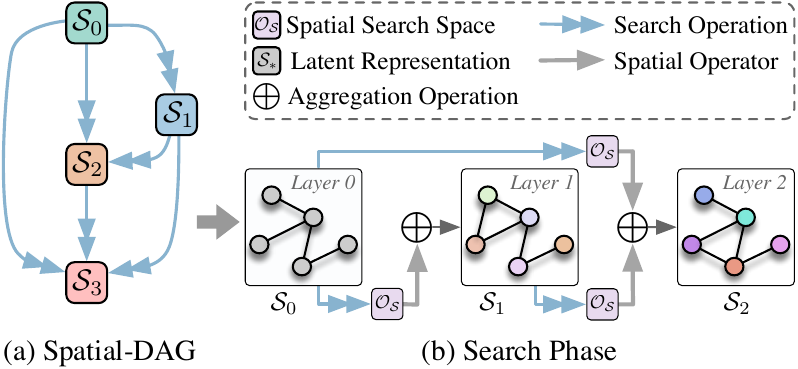}
  \vspace{-0.2cm}
  \caption{The spatial search in a Spatial-DAG. $\mathcal{S}_*$ denote the latent representation. 
  The search operation aims to identify an optimal spatial operator between any two nodes (e.g., $\mathcal{S}_1 \to \mathcal{S}_2$). Once identified, this operator transfers the latent representation from $\mathcal{S}_1$ to $\mathcal{S}_2$.}
  \label{fig:spatial_dag}
  \Description{}
  \vspace{-0.2cm}
\end{figure}

\subsection{Spatial Search Module}
\label{sec:spatial_search}
To model fine-grained spatio-temporal dependencies for each patch embedding $h_{\mathcal{S}_m}$, we design a new spatial search space to cover a wide range of spatial modeling paradigms and automatically optimize messages-passing across different layers, leading to more accurate forecasting.
Figure~\ref{fig:spatial_dag} is the detailed framework of the spatial search module.

\textbf{Spatial search space.}
After analyzing existing literature (refer to Section~\ref{sec:The Effectiveness Bottleneck}), we classify the adjacency matrix into three categories: fully fixed, semi-adaptive, and fully adaptive (see table \ref{tab:optimal_spatial_operation}). 
To enhance the flexibility and capacity of exploring spatio-temporal dependencies, we also incorporate $Zero$ and $Identity$ into the spatial search space.
We define the spatial search space as $\mathcal{O}_{\mathcal{S}}$ = $\{GNN_{fixed}$, $GNN_{adap}$, $GNN_{att}$, $Zero, Identity\}$, where $GNN_{fixed}$, $GNN_{adap}$, and $GNN_{att}$ denote fully fixed, semi-adaptive, and fully adaptive matrix used in GNN layer, respectively.

\textbf{$GNN_{fixed}$.}
This operator employs a distance-based matrix for message-passing aggregation and is the predefined adjacency matrix, fixed in both the training and inference phases.
The formal formulation is as follows:
\begin{equation}
\begin{aligned}
GNN_{fixed}(X) = \sum_{k=0}^{K} \mathbb{A}_{dis}^{k}XW_{fixed},
\end{aligned}
\end{equation}
where $\mathbb{A}_{dis}$ is the distance matrix and $k$ is the finite steps. $W_{fixed}$ is the learnable weights and $X$ is the input of $GNN_{fixed}$ operator.

\textbf{$GNN_{adap}$.}
This operator utilizes an adaptive matrix that is learned from node embedding. 
We refer to $GNN_{adap}$ as a semi-adaptive matrix because the adjacency matrix in this operator is trainable during the training phase but remains fixed during the inference phase.
The adaptive matrix has a strong ability to model spatial dependencies and is commonly used in many spatio-temporal forecasting models~\cite{wu2019graph,bai2020adaptive,wu2020connecting,zhang2020spatio}. 
As is common in many models, we randomly generate two sets of node embeddings $E_1 \in \mathbb{R}^{N\times d}$ and $E_2 \in \mathbb{R}^{N\times d}$ that are represented by learnable weights. We then compute the adaptive matrix using these two node embeddings:
$\mathbb{A}_{adap} = SoftMax(ReLU(E_{1} E_{2}^{T}))$.
We use this adaptive matrix to tackle the massage-passing aggregation of the information for their neighborhood. The formal formulation is as follows:
\begin{equation}
\begin{aligned}
GNN_{adap}(X) = \sum_{k=0}^{K} \mathbb{A}_{adap}^{k}XW_{adap},
\end{aligned}
\end{equation}
where $W_{adap}$ is learnable weights and $X$ is the input of this operator.

\textbf{$GNN_{att}$.}
Attentions are utilized in many state-of-the-art spatio-temporal forecasting models. 
This operator utilizes multi-head attention to compute a fully adaptive matrix based on the input time series. 
Notably, this fully adaptive matrix in operator $GNN_{att}$ is automatically generated during the training and inference phases.

We briefly introduce the multi-attention mechanism, which is a concatenation of $t$ self-attention heads. 
Specifically, given attention head $t$, the learnable matrices $W_i^Q, W_i^K, W_i^V \in \mathbb{R}^{D\times \frac{D}{t}}$, the attention function can be written as:
\begin{equation}
\begin{aligned}
{\mathbf{Q}}_{i} = XW_i^Q, \mathbf{K}_{i} = XW_i^K, \mathbf{V}_{i} = XW_i^V,
\end{aligned}
\end{equation}
\begin{equation}
\begin{aligned}
\mathbb{A}_{att} = softmax(\frac{\mathbf{Q}_{i} \cdot \mathbf{K}_{i}^T}{\sqrt{D/t}}),
\end{aligned}
\end{equation}
where $\mathbb{A}_{att}$ represents the fully adaptive matrix. $D$ denotes the number of hidden dimensions. Then, we can compute the message-passing aggregation from neighboring nodes:
\begin{equation}
\begin{aligned}
\label{eq:att}
h^{\prime} = concat(\{ \mathbb{A}_{att} \cdot \mathbf{V}_i \}_{i=1}^{t}).
\end{aligned}
\end{equation}
After this, the embedding $h^{\prime}$ is fed into linear layers, which outputs the final embedding calculated by this operator. 
The formal formulation is as follows:
\begin{equation}
\begin{aligned}
GNN_{att}(X) = (ReLU(h^{\prime}\cdot W_1 + b_1)) \cdot W_2 + b_2.
\end{aligned}
\end{equation}

Drawing inspiration from \cite{lai2023lightcts}, we also adopt a grouping strategy in the $GNN_{att}$ operator. 
The input $X$ of this operator is initially partitioned into several groups, and Equation \ref{eq:att} is subsequently applied to each group, which can enhance computational efficiency.

\textbf{Spatial-DAG.}
To explore the fine-grained temporal and spatial dependencies (refer to Figure \ref{fig:multi-scale}), we construct M spatial-DAGs for each patch embedding (\eg $h_{\mathcal{S}_1}$) in the $H_\mathcal{P}$.
In each spatial-DAG, spatio-temporal dependencies are explored by automatically selecting the optimal message-passing aggregation in different layers. 
It is important to note that in order to address the issue of parameter explosion, the network parameters of spatial operators on the same type edge (\eg the edge between $\mathcal{S}_0$ and $\mathcal{S}_1$) are shared among different spatial-DAGs. 
However, the architecture parameters $\beta$ on different spatial-DAG are trained separately. 
In other words, optimal spatial operators can be automatically selected for different spatial-DAGs.

\textbf{Parameterizing spatial-DAG.}
To provide a clearer explanation, we consider the parameterizing process of one spatial-DAG for one patch embedding as an example.
Similar to the temporal-DAG, we denote $\beta$ as the architecture parameters of the spatial-DAG. 
$\beta^{(i\rightarrow j)}_{o}$ is the weight of spatial operator $o$ (\eg $GNN_{adap}$) from the node $\mathcal{S}_i$ and $\mathcal{S}_j$ in the spatial-DAG.
For the patch embeddings $h_{\mathcal{S}_m}$, the formulation of parameterizing this spatial-DAG is as follows:
\begin{equation}
\begin{aligned}
h^m_{\mathcal{S}} = \sum_{j=1}^{N_{\mathcal{S}}} \sum_{i=0}^{i<j} \sum_{o\in \mathcal{O}_{\mathcal{S}}} \frac{exp(\beta^{(i\rightarrow j)}_{o})}{\sum_{o^\prime \in \mathcal{O}_{\mathcal{S}}} exp(\beta{o^\prime}^{(i\rightarrow j)})} o(\mathcal{S}_i),
\end{aligned}
\end{equation}
where $\mathcal{S}_0 = h_{\mathcal{S}_m}$ server as the input of this spatial-DAG. $N_\mathcal{S}$ is the number of nodes in this spatial-DAG, and , $h^m_{\mathcal{S}}$ denotes the output embedding of this $m$-th spatial-DAG.

Finally, using the same method, all the patch embeddings can be calculated to obtain the output of each spatial-DAG, denoted as $h_{\mathcal{S}} = \{ h^1_{\mathcal{S}}, \cdots, h^m_{\mathcal{S}}, \cdots, h^M_{\mathcal{S}} \}$.
We sum the outputs of each spatial-DAG to obtain the final output of the spatial search module, the formulation is as follows:
\begin{equation}
\begin{aligned}
H_{\mathcal{S}} = \sum_{m=0}^{M} h^m_{\mathcal{S}},
\end{aligned}
\end{equation}
where $H_{\mathcal{S}}$ denotes the output of spatial search module.

\subsection{Output Layer and Search Strategy}
\label{sec:search_strategy}
In the output layer, we concatenate the skip embedding as the final representation: $H^{\prime} = E_{emb} \oplus H_{\mathcal{T}} \oplus H_{\mathcal{S}}$, 
where $\oplus$ denotes the concatenation operation.
Then, the final presentation is fed into the linear layers to make the spatio-temporal forecasting.

In our AutoSTF, there are mainly two types of parameters: architecture parameters, such as weights of candidate operators, and network parameters, such as internal parameters of spatial and temporal operators.
We denote the architecture parameters and network weight parameters as $\Theta = \{\alpha, \beta \}$ and $\omega$, respectively, and all computations are differentiable.
Therefore, we can employ a bi-level gradient-based optimization algorithm like DARTS \cite{liu2018darts}, which is a gradient-based neural network architecture search algorithm.:
\begin{equation}
\begin{aligned}
\underset{\alpha}{\min} \mathcal{L}_{val}(\omega^{*}, \Theta), \\
\st \omega^{*} = \arg \underset{\omega}{\min} \mathcal{L}_{train}(\omega, \Theta),
\end{aligned}
\end{equation}
where $\mathcal{L}_{train}$ and $\mathcal{L}_{val}$ denote the loss value on the training dataset and validation dataset, respectively.
In other words, we iteratively update weight parameters using the training dataset $\mathcal{D}_{train}$ and architecture parameters using the validation dataset $\mathcal{D}_{val}$. 


\section{Experiments}
\label{sec:Experiments}

In this section, we conducted experiments on both multi- and single-step spatio-temporal forecasting to evaluate AutoSTF. 
We introduce the experimental settings, overall results of AutoSTF, efficiency study, and ablation study. 
Due to page limitations, we have provided the parameter sensitivity analysis and case study in Appendices \ref{appdix:Parameter Sensitivity Analysis} and \ref{appdix:Case Study}, respectively. In Appendix \ref{appdix:The Improvement of AutoSTF}, we include a detailed figure to demonstrate the significant improvements of our model compared to other automated models. We analyze the algorithmic complexity of AutoSTF in Appendix \ref{appdix:Complexity Analysis}.
In addition, we conducted a comprehensive analysis and extensive experiments of decoupled the search space in Appendices \ref{appdix:The Analysis of Decoupled Search Space}, which involved comparing the model's performance and efficiency before and after decoupling, as well as its ability to maintain spatiotemporal correlations. 

%

\subsection{Experimental Settings}
In this subsection, we describe the datasets used, the metrics employed in our experiments, and the baseline comparisons we have conducted. 
Detailed information on hyper-parameters and implementation specifics can be found in Appendix \ref{appendix:Implementation Details}.

\textbf{Datasets.}
We performed experiments on eight benchmark datasets to evaluate both multi-step and single-step forecasting. For multi-step forecasting, we utilized traffic speed datasets (METR-LA and PEMS-BAY) provided by Li et al. \cite{li2017diffusion} and traffic flow datasets (PEMS03, PEMS04, PEMS07, and PEMS08) released by Song et al. \cite{song2020spatial}. As for single-step forecasting, we employed solar-energy and electricity datasets made available by Lai et al. \cite{lai2018modeling}. 
These datasets can be categorized as traffic speed, traffic flow, solar-energy, and electricity.
We provide the detailed statistics of the datasets in Appendix \ref{appendix:Datasets}.

\textbf{Evaluation Metrics.}
We followed previous works \cite{wu2021autocts,pan2021autostg,wu2023autocts+} to evaluate the model's performance in multi- and single-step forecasting. 
Specifically, Mean Absolute Error (MAE), Root Mean Squared Error (RMSE), and Mean Absolute Percentage Error (MAPE) were utilized to assess the accuracy of multi-step forecasting. 
For single-step forecasting, we utilized Root Relative Squared Error (RRSE) and Empirical Correlation Coefficient (CORR) as evaluation metrics. 
Lower values of MAE, RMSE, MAPE, and RRSE are indicative of higher accuracy, while larger CORR values denote higher accuracy.
More details are stated in Appendix \ref{appendix:Evaluation Metrics}.

\begin{table}[b] 
\vspace{-0cm}
\centering
\footnotesize
\setlength{\tabcolsep}{2.5pt} 

\caption{The performance of multi-step forecasting on traffic speed datasets.}
\label{tab:traffic_speed}
\begin{tabular}{c|c|ccc|ccc|ccc}
\toprule
\multicolumn{2}{c|}{\multirow{2}{*}{Models}} & \multicolumn{3}{c|}{15 min} & \multicolumn{3}{c|}{30 min} & \multicolumn{3}{c}{60 min} \\
\multicolumn{2}{c|}{} & MAE & RMSE & MAPE & MAE & RMSE & MAPE & MAE & RMSE & MAPE \\
\midrule
\multirow{12}{*}{\rotatebox{90}{METR-LA}} 
& DCRNN         & 2.77 & 5.38 & 7.30\% & 3.15 & 6.45 & 8.80\% & 3.60 & 7.60 & 10.50\% \\
& STGCN         & 2.88 & 5.74 & 7.62\% & 3.47 & 7.24 & 9.57\% & 4.59 & 9.40 & 12.70\% \\
& GWN           & 2.69 & 5.15 & 6.90\% & 3.07 & 6.22 & 8.37\% & 3.53 & 7.37 & 10.01\% \\
& AGCRN         & 2.83 & 5.45 & 7.56\% & 3.20 & 6.55 & 8.79\% & 3.58 & 7.41 & 10.13\% \\
& MTGNN         & 2.69 & 5.18 & 6.86\% & 3.05 & 6.17 & 8.19\% & 3.49 & 7.23 & 9.87\%  \\
& STID          & 2.82 & 5.53 & 7.75\% & 3.19 & 6.57 & 9.39\% & 3.55 & 7.55 & 10.95\%  \\
& STAEformer    & 2.74 & 5.33 & 7.11\% & 3.07 & 6.28 & 8.42\% & 3.46 & 7.28 & 9.97\%  \\
& LightCTS      & \underline{2.67} & 5.16 & \underline{6.82\%} & \underline{3.03} & 6.16 & \underline{8.11\%} & \underline{3.42} & 7.21 & \textbf{9.46\%}  \\
& AutoSTG       & 2.70 & 5.16 & 6.91\% & 3.06 & 6.17 & 8.30\% & 3.47 & 7.27 & 9.87\%  \\
& AutoCTS       & \underline{2.67} & \underline{5.11} & \textbf{6.80\%} & 3.05 & \underline{6.11} & 8.15\% & 3.47 & \underline{7.14} & 9.81\%  \\
& AutoCTS+       & \underline{2.67} & 5.18 & 6.83\% & 3.06 & 6.15 & 8.19\% & \underline{3.42} & 7.16 & 9.79\%  \\
& \textbf{AutoSTF} & \textbf{2.65} & \textbf{5.10} & \textbf{6.80\%} & \textbf{2.99} & \textbf{6.10} & \textbf{8.10\%} & \textbf{3.35} & \textbf{7.09} & \underline{9.58\%}  \\
\midrule
\multirow{12}{*}{\rotatebox{90}{PEMS-BAY}}
& DCRNN         & 1.38 & 2.95 & 2.90\% & 1.74 & 3.97 & 3.90\% & 2.07 & 4.74 & 4.90\% \\
& STGCN         & 1.36 & 2.96 & 2.90\% & 1.81 & 4.27 & 4.17\% & 2.49 & 5.69 & 5.79\% \\
& GWN           & 1.30 & 2.74 & 2.73\% & 1.63 & 3.70 & 3.67\% & 1.95 & 4.52 & 4.63\% \\
& AGCRN         & 1.35 & 2.83 & 2.87\% & 1.69 & 3.81 & 3.84\% & 1.96 & 4.52 & 4.67\% \\
& MTGNN         & 1.32 & 2.79 & 2.77\% & 1.65 & 3.74 & 3.69\% & 1.94 & 4.49 & 4.53\% \\
& STID          & 1.31 & 2.79 & 2.78\% & 1.64 & 3.73 & 3.73\% & 1.91 & 4.42 & 4.55\%  \\
& STAEformer    & 1.31 & 2.78 & 2.76\% & 1.62 & 3.68 & 3.62\% & \underline{1.88} & 4.34 & 4.41\%  \\
& LightCTS      & \underline{1.30} & 2.75 & 2.71\% & \underline{1.61} & 3.65 & 3.59\% & 1.89 & 4.32 & 4.39\% \\
& AutoSTG       & 1.31 & 2.76 & 2.73\% & 1.63 & 3.67 & 3.63\% & 1.92 & 4.38 & 4.43\% \\
& AutoCTS       & \underline{1.30} & \underline{2.71} & \underline{2.69\%} & \underline{1.61} & \underline{3.62} & \underline{3.55\%} & 1.89 & 4.32 & 4.36\% \\
& AutoCTS+       & 1.31 & 2.73 & \underline{2.69\%} & 1.63 & 3.63 & 3.61\% & \underline{1.88} & \underline{4.30} & \underline{4.34\%}  \\
& \textbf{AutoSTF} & \textbf{1.27} & \textbf{2.70} & \textbf{2.64\%} & \textbf{1.56} & \textbf{3.59} & \textbf{3.48\%} & \textbf{1.81} & \textbf{4.25} & \textbf{4.23\%} \\
\bottomrule
\end{tabular}
\end{table}

\begin{table*}[t] \footnotesize
  \setlength{\tabcolsep}{2.5pt} 
  \caption{The performance of multi-step forecasting on traffic flow datasets.}
  \label{tab:traffic_flow}
  \begin{tabular}{c|c|ccccccccccc|cccc}
    \toprule

    Data & Metric & DCRNN & STGCN & GWN & AGCRN & MTGNN & STID & STHODE & DeepSTUQ & STJGCN & STAEformer & LightCTS & AutoSTG & AutoCTS & AutoCTS+ &  \textbf{AutoSTF} \\
    
    \midrule
    \multirow{3}{*}{\rotatebox{90}{PEMS03}}
    & \text{ MAE }  & 18.18   & 17.49   & 14.82   & 15.89   & 15.10   & 15.49   & 15.51   & 15.13     & 14.92    & 14.96   & 15.23   & 17.97   & 14.71   & \underline{14.59}   & \textbf{14.44}\\
    & \text{ RMSE } & 30.31   & 30.12   & 25.24   & 28.12   & 25.93   & 28.43   & 26.16   & 26.77     & 25.70    & 26.03   & 26.21   & 28.47   & 24.54   & \underline{24.21}   & \textbf{23.94}\\
    & \text{ MAPE } & 18.91\% & 17.15\% & 16.16\% & 15.38\% & 15.67\% & 16.03\%   & 15.88\%   & 14.03\% & 14.81\%  & 15.06\%   & 14.81\% & 18.08\% & 14.39\% & \underline{14.02\%} & \textbf{13.79}\%\\

    \midrule
    \multirow{3}{*}{\rotatebox{90}{PEMS04}}
    & \text{ MAE }  & 24.70   & 22.70   & 19.16   & 19.83   & 19.32   & 19.25   & 19.61   & 19.11     & 18.81     & \underline{18.41}   & 18.79    & 20.46   & 19.13   & 18.95   & \textbf{18.38}\\
    & \text{ RMSE } & 38.12   & 35.55   & 30.46   & 32.26   & 31.57   & 30.91   & 30.97   & 31.68     & 30.35     & \underline{30.01}   & 30.14    & 32.18   & 30.44   & 30.31   & \textbf{29.86}\\
    & \text{ MAPE } & 17.12\% & 14.59\% & 13.26\% & 12.97\% & 13.52\% & 13.30   & 13.45\%   & 12.71\% & \textbf{11.92\%}   & 13.36\%   & 12.80\%  & 13.77\% & 12.89\% & 12.75\% & \underline{12.58\%}\\

    \midrule
    \multirow{3}{*}{\rotatebox{90}{PEMS07}}
    & \text{ MAE }  & 25.30   & 25.38   & 21.54  & 21.31  & 22.07   & 20.69   & 21.71    & 20.36   & \underline{19.95}    & 20.64   & 23.24   & 26.77    & 20.93   & 20.77   & \textbf{19.50}\\
    & \text{ RMSE } & 38.58   & 38.78   & 34.23  & 35.06  & 35.80   & 33.93   & 34.63    & 33.71   & \underline{33.01}    & 33.53   & 36.86   & 41.63    & 33.69   & 33.49   & \textbf{32.66}\\
    & \text{ MAPE } & 11.66\% & 11.08\% & 9.22\% & 9.13\% & 9.21\%  & 8.82\%   & 9.82\%   & 8.63\%  & \underline{8.31\%}   & 8.56\%   & 9.87\% & 11.63\%  & 8.90\% & 8.76\%  & \textbf{8.14\%}\\

    \midrule
    \multirow{3}{*}{\rotatebox{90}{PEMS08}}
    & \text{ MAE }  & 17.86   & 18.02   & 15.13   & 15.95   & 15.71   & 15.28   & 15.43     & 15.44     & \underline{14.53}    & 14.78   & 14.63  & 16.23    & 14.82   & 14.72   & \textbf{14.07}\\
    & \text{ RMSE } & 27.83   & 27.83   & 24.07   & 25.22   & 24.62   & 24.85   & 24.39     & 24.60     & 23.73    & 23.97   & 23.49  & 25.72    & 23.64   & \underline{23.43}   & \textbf{23.17}\\
    & \text{ MAPE } & 11.45\% & 11.40\% & 10.10\% & 10.09\% & 10.03\% & 9.92\%   & 10.27\%   & 10.06\%   & \underline{9.15\%}   & 9.71\%   & 9.43\% & 10.25\%  & 9.51\%  & 9.45\%  & \textbf{9.14\%}\\
    
    \bottomrule
  \end{tabular}
\end{table*}

\textbf{Baselines.}
We compare AutoSTF with 14 manual-designed methods, including DCRNN~\cite{li2017diffusion}, STGCN~\cite{yu2018spatio}, Graph WaveNet~\cite{wu2019graph}, AGCRN~\cite{bai2020adaptive}, MTGNN~\cite{wu2020connecting}, STID~\cite{shao2022spatial}, STHODE~\cite{yao2023spatio}, DeepSTUQ~\cite{qian2023uncertainty}, STJGCN~\cite{zheng2023spatio}, STAEformer~\cite{liu2023spatio} and LightCTS~\cite{lai2023lightcts}, and 3 automated models including AutoSTG \cite{pan2021autostg}, AutoCTS+~\cite{wu2023autocts+} and AutoCTS~\cite{wu2021autocts}.
We provide the detailed description of baselines in Appendix \ref{appendix:Baselines}.


\subsection{Overall Results of AutoSTF}
Firstly, we present the main experiments for both multi-step (refer to Table \ref{tab:traffic_speed} and \ref{tab:traffic_flow}) and single-step (see Table \ref{tab:single_step}) spatio-temporal forecasting.
We adopt the convention of highlighting the best accuracy \textbf{in bold} and \underline{underlining} the second best accuracy.

\textbf{Multi-step forecasting.}
As demonstrated in Tables \ref{tab:traffic_speed} and \ref{tab:traffic_flow}, AutoSTF outperforms all baseline models on all multi-step forecasting tasks across all evaluation metrics. 
In summary, the results demonstrate that AutoSTF can produce highly competitive architectures adept at capturing spatio-temporal dependencies, leading to enhanced forecasting accuracy.

In particular, compared with the most advanced manual-designed baseline LightCTS, our model achieved significant improvements on all datasets. 
For example, our model resulted in a (2.05\%, 1.32\%, 0.76\%) and (4.23\%, 3.11\%, 2.31\%) reduction in MAE for long-term (60 min), medium-term (30 min), and short-term (15 min) forecasting on METR-LA and PEMS-BAY datasets, respectively. 
Furthermore, our model demonstrated enhanced performance on other traffic flow datasets compared to LightCTS, achieving a reduction in MAE (2.18\%), RMSE (0.93\%), and MAPE (1.72\%) for PEMS04, as well as a decrease in the same metrics for PEMS08, with MAE (3.83\%), RMSE (1.36\%), and MAPE (3.08\%).
Two crucial insights can be drawn from our observations.
Firstly, the experimental outcomes reveal that AutoSTF is capable of automatically generating highly competitive neural architectures that surpass human-designed models in performance.
secondly, no hand-designed model consistently outperforms others across various traffic flow datasets. 
This implies that different datasets require distinct model architectures. 
Nonetheless, our AutoSTF can adapt to these requirements by automatically designing optimal neural architectures for different traffic flow datasets, ultimately yielding the most accurate forecasting results.

Moreover, when compared to the automated models, our proposed AutoSTF demonstrates significant advancements across all datasets in terms of multi-step forecasting performance.
Specifically, compared with AutoCTS on long-term forecasting, our AutoSTF results in a (3.46\%, 0.70\%, 2.34\%), (4.23\%, 1.65\%, 2.98\%), (3.92\%, 1.91\%, 2.40\%), and (5.06\%, 1.99\%, 3.89\%) improvement on the METR-LA, PEMS-BAY, PEMS04, and PEMS08 datasets, respectively, under the MAE, RMSE, and MAPE metrics.
We can draw two key observations as follows:
Firstly, compared to other automated models, our proposed AutoSTF achieves superior performance on both traffic speed and traffic flow datasets, demonstrating the effectiveness of our framework in accurately capturing spatio-temporal dependencies.
Secondly, the experimental results also indicate that fine-grained spatial search and adaptive selection of the optimal message-passing aggregation in different layers can significantly improve accuracy.

\begin{table}[t]\footnotesize
  \caption{The performance of single-step forecasting.}
  \label{tab:single_step}
  \begin{tabular}{c|c|cc|cc}
    \toprule

    \multirow{2}{*}{\text { Data }} & \multirow{2}{*}{\text { Models }} & \multicolumn{2}{c|}{3-th} & \multicolumn{2}{c}{24-th}\\

    & & \text { RRSE $\downarrow$} & \text { CORR $\uparrow$} & \text { RRSE $\downarrow$ } & \text { CORR $\uparrow$} \\
    
    \midrule

    \multirow{7}{*}{\text { \rotatebox{0}{Solar-energy} }}
    & \text{ MTGNN }         & 0.1778 & 0.9852 & 0.4270 & 0.9031 \\
    & \text{ AGCRN }         & 0.1830 & 0.9846 & 0.4602 & 0.9016 \\
    & \text{ LightCTS }      & 0.1714 & 0.9864 & 0.4129 & 0.9084 \\
    & \text{ AutoSTG }       & 0.2094 & 0.9811 & 0.5066 & 0.8611 \\
    & \text{ AutoCTS }       & 0.1750 & 0.9853 & 0.4143 & 0.9085 \\
    & \text{ AutoCTS+ }      & \textbf{0.1657} & \underline{0.9875} & \underline{0.3980} & \underline{0.9152} \\
    & \textbf{ AutoSTF }       & \underline{0.1682} & \textbf{0.9882} & \textbf{0.3957} & \textbf{0.9175} \\
    
    \midrule
    \multirow{7}{*}{\text { \rotatebox{0}{Electricity}}}
    & \text{ MTGNN }         & 0.0745 & 0.9474 & 0.0953 & 0.9234 \\
    & \text{ AGCRN }         & 0.1033 & 0.8854 & 0.0994 & 0.9073 \\
    & \text{ LightCTS }      & 0.0736 & 0.9445 & 0.0952 & 0.9215 \\
    & \text{ AutoSTG }       & 0.1188 & 0.9070 & 0.0998 & 0.8846 \\
    & \text{ AutoCTS }       & 0.0743 & 0.9477 & 0.0947 & 0.9239 \\
    & \text{ AutoCTS+ }      & \underline{0.0736} & \underline{0.9483} & \textbf{0.0921} & \underline{0.9253} \\
    & \textbf{ AutoSTF }       & \textbf{0.0732} & \textbf{0.9497} & \underline{0.0935} & \textbf{0.9279} \\

    \bottomrule
  \end{tabular}
  \vspace{-0.2cm}
\end{table}

\textbf{Single-step forecasting.}
In this experiment, we compared AutoSTF with other automated spatio-temporal forecasting models and presented the results in Table \ref{tab:single_step} for the single-step forecasting task. 
Consistent with prior research \cite{wu2021autocts,wu2023autocts+}, we present the single-step forecasting results for the 3rd and 24th future time steps.
It should be noted that AutoSTG utilizes a predefined adjacency matrix for its spatial graph convolution operators. 
However, since this matrix is not available in single-step spatio-temporal datasets, the spatial graph convolution operators can be excluded from AutoSTG's search space when running it on such datasets. 

Overall, AutoSTF outperformed the other models and achieved state-of-the-art results on the Solar-Energy and Electricity datasets for almost all forecasting horizons. 
Notably, AutoSTF demonstrated a significant improvement in the CORR metric and outperformed all other models on these datasets. 
The improvements achieved by AutoSTF on different datasets highlight the importance of automated solutions in identifying specific and optimal models.

\begin{figure}[t]
  \centering
  \includegraphics[width=1\linewidth]{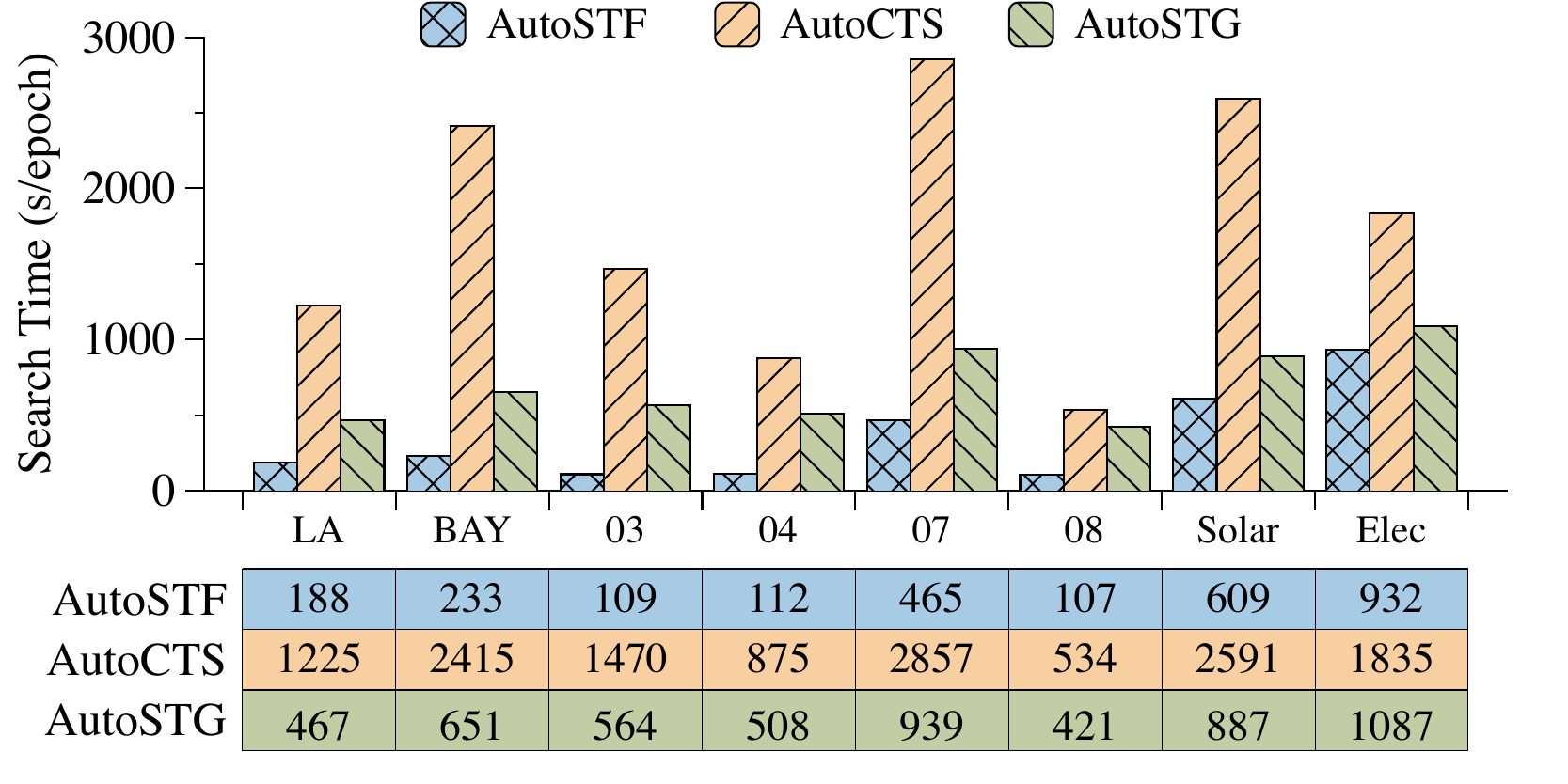}
  \vspace{-0.3cm}
  \caption{The search time of all the datasets.}
  \label{fig:efficiency_time}
  \Description{}
\end{figure}

\begin{figure}[t]
  \centering
  \includegraphics[width=1\linewidth]{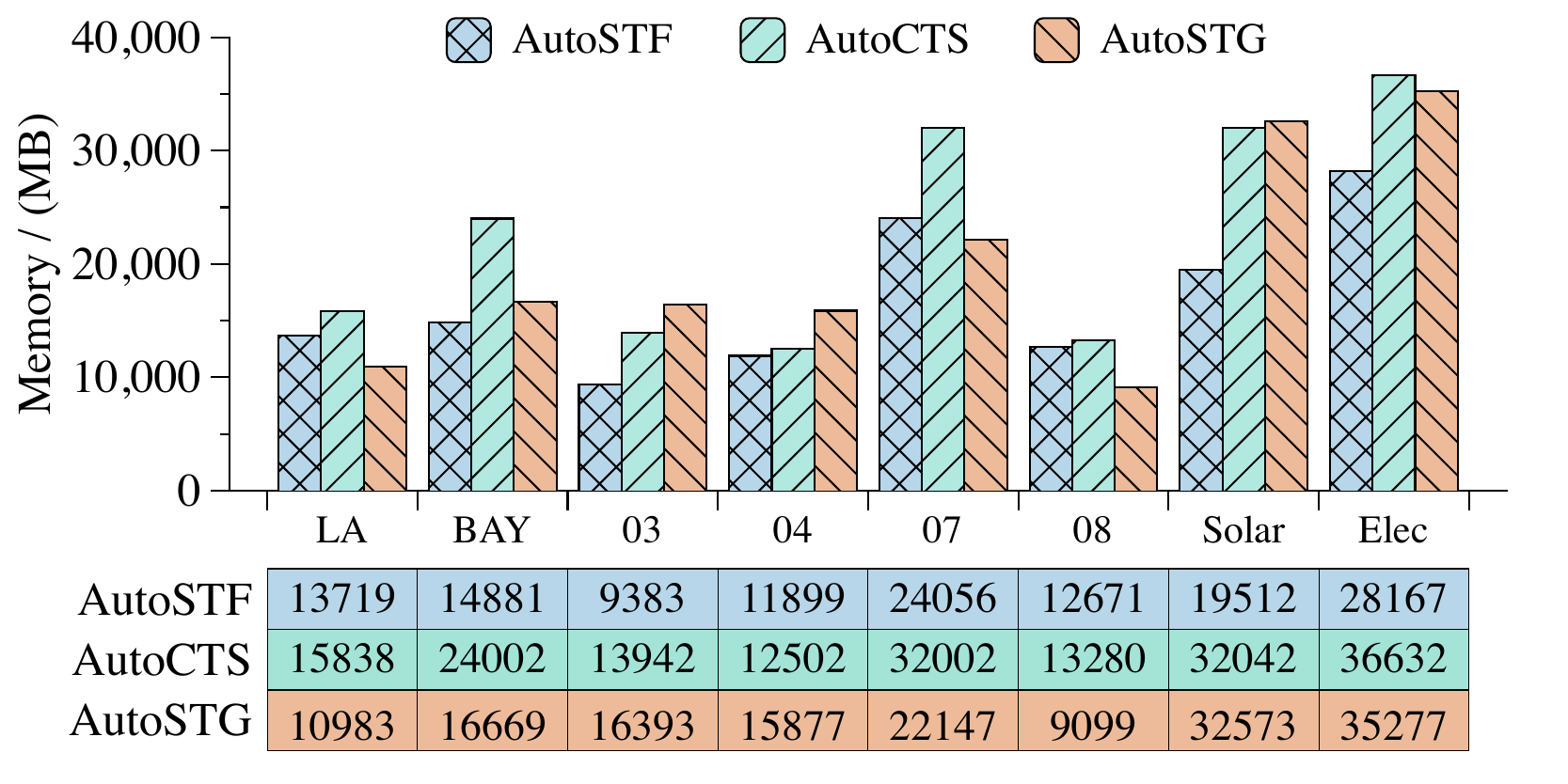}
  \vspace{-0.3cm}
  \caption{The memory consumption in the search phase.}
  \label{fig:efficiency_memory}
  \Description{}
\end{figure}

\subsection{Efficiency Study}

For efficiency, in this study, we place significant emphasis on optimizing the search efficiency. 
We selected AutoCTS and AutoSTG, which are among the most advanced automated spatio-temporal forecasting models, as baselines for comparing the search time and memory consumption in the search phase.
It is important to note that the experimental setup for AutoCTS+ \cite{wu2023autocts+} is significantly different from that of AutoSTF, AutoCTS, and AutoSTG, as it involves training an Architecture-Hyperparameter Comparator using a large volume of samples to predict the final neural architecture. 
Therefore, to ensure an accurate comparison, we excluded AutoCTS+ from the efficiency evaluation. The efficiency experiments were conducted on a Linux Ubuntu server equipped with 2 V100 GPUs.

Figures \ref{fig:efficiency_time} and \ref{fig:efficiency_memory} present a comparison of the average search time per epoch and the memory consumption across different datasets, respectively.
Our AutoSTF consistently demonstrates the best performance in terms of search time across all datasets, highlighting the efficiency of our proposed framework.
From the results, we can draw two key insights as follows.
Firstly, AutoCTS takes thousands of seconds to complete an epoch search on most datasets, resulting in excessive computational overhead and presenting significant challenges.
In contrast, AutoSTF only takes hundreds of seconds to complete the one epoch search, significantly improving computational efficiency.
This demonstrates that decoupling the mixed search space, reducing redundant temporal features and parameter-sharing schemes for spatial search can effectively enhance computational efficiency.
Secondly, regarding memory computations, AutoSTF exhibits the smallest consumption on most datasets, indicating its superior performance in terms of memory efficiency.
In conclusion. the experimental results clearly demonstrate that our model is significantly more efficient than AutoCTS and AutoSTG in terms of search time and memory consumption across all datasets.

\begin{table}[t]\footnotesize
  \caption{Ablation studies on different type datasets.}
  \label{tab:ablation_studies}
  \begin{tabular}{c|c|ccc|c}
    \toprule

    \text { Dataset } & \text { Models } & \text { MAE } & \text { RMSE } & \text { MAPE } & \text { Search Time / (s) }\\

    \midrule
    \multirow{4}{*}{\rotatebox{90}{METR-LA}}
    & \text{ w/o TS }      & 3.39 & 7.19 & 9.65   & 58.40 \\
    & \text{ w/o MPT }     & 3.38 & 7.19 & 9.85   & 152.12 \\
    & \text{ w/o SS }      & 3.53 & 7.58 & 10.66  & 95.89 \\
    & \textbf{ AutoSTF }     & \textbf{3.35} & \textbf{7.09} & \textbf{9.58}  & 188 \\

    \midrule
    \multirow{4}{*}{\rotatebox{90}{PEMS04}}
    & \text{ w/o TS }      & 18.62 & 30.09 & 12.81 & 34.46 \\
    & \text{ w/o MPT }     & 18.67 & 30.14 & 12.86 & 87.93 \\
    & \text{ w/o SS }      & 18.58 & 30.06 & 12.71 & 58.96 \\
    & \textbf{ AutoSTF }     & \textbf{18.38} & \textbf{29.86} & \textbf{12.58} & 112 \\
    \bottomrule
  \end{tabular}
\end{table}

\subsection{Ablation Study}

To investigate the contribution of each key module to improving the performance and efficiency of the search phase of the proposed AutoSTF model, we performed an ablation study on three variants of AutoSTF. 
The three variants are described as follows:
(1) "w/o TS" refers to the model variant that omits the Temporal Search module, instead utilizing a Linear layer as its replacement.
(2) "w/o MPT" is the model variant that removed the Multi-Patch Transfer module, which means this model variant only contains a temporal-DAG and a spatial-DAG.
(3) "w/o SS" is the model variant that removed the Spatial Search module.
We present the experimental results in Table \ref{tab:ablation_studies} and provide a discussion of the findings.

Table \ref{tab:ablation_studies} demonstrates that AutoSTF outperforms its variants, each of which disregards a different component of AutoSTF. 
Specifically, AutoSTF achieves higher accuracy compared to w/o TS and w/o SS, which highlights the effectiveness that AutoSTF achieved by decoupling the mixed search space into temporal and spatial space.
Furthermore, AutoSTF significantly outperforms the w/o MPT variant, reinforcing the effectiveness of multi-patch transfer in improving accuracy by incorporating temporal information to model finer-grained spatio-temporal dependencies.
In terms of search time, it can be concluded that the temporal search is the most time-consuming part. This can be attributed to the inclusion of the informer as the temporal operator in the temporal search, which is a particularly computationally demanding model.
Furthermore, it should be noted that the search time of the w/o MPT variant is lower than that of AutoSTF. This is because when the MPT module is removed, the temporal search embedding is not split into multiple patches. As a result, there is only one spatial-DAG to explore the spatio-temporal correlation, leading to a reduced search time.

\section{Related Work}
\label{sec:Related Work}
This section emphasizes related work in two aspects: manually-designed and automated models in spatio-temporal forecasting.

\textbf{The manually-designed models.}
Existing manually-designed models often propose complex algorithms to model the spatial-temporal dependencies using domain-specific knowledge \cite{liu2023spatio,zheng2023spatio,qian2023uncertainty,yao2023spatio}. 
In particular, many works~\cite{li2017diffusion,pan2019urban,yu2018spatio,li2023sastgcn,shao2022pre,jiang2023spatio} involve Spatial-Temporal Graph Neural Networks (STGNN) to explore the temporal dependencies within time series and spatial correlations across different locations for spatio-temporal forecasting.
For temporal modeling, techniques such as Gated Recurrent Units (GRU), Long Short-Term Memory (LSTM)~\cite{sutskever2014sequence}, and Temporal Convolutional Networks (TCN)~\cite{yu2015multi} are commonly employed to extract information from time series data. 
In addition, various Graph Neural Network approaches have been proposed for modeling spatial dependencies using different adjacency matrices, including predefined and adaptive matrices~\cite{wu2019graph,li2017diffusion}.
Graph WaveNet~\cite{wu2019graph}, as proposed by Wu et al., utilizes both predefined and adaptive matrices to capture spatio-temporal dependencies.
Furthermore, several studies have proposed methods to model dynamic spatial dependencies~\cite{han2021dynamic,shao2022decoupled,zhang2022dynamic,ye2022learning}. 
These methods primarily focus on learning the hidden relationships between nodes using dynamic node features, which are generally characterized by a combination of real-time traffic attributes.
For example, ESG~\cite{ye2022learning} presents a flexible graph neural network that models multi-scale time series interactions to simultaneously capture pairwise correlations and temporal dependencies.
However, manually-designed models necessitate substantial domain-specific expertise, as well as extensive parameter adjustments.

\textbf{The automated models.}
Recently, there has been a growing trend toward developing automated methods for designing efficient model architectures, aiming to enhance the accuracy of forecasting predictions \cite{li2020autost,zhang2023autostl,ke2023autostg+}.
The existing automated time series forecasting models aim to search the optimal architecture in the mixed search space for modeling the spatio-temporal dependencies. 
For instance, AutoSTG~\cite{pan2021autostg} employs an adaptive matrix learned from node features using meta-learning and explores spatio-temporal dependencies within a mixed search space.
AutoCTS~\cite{wu2021autocts} employs micro and macro search strategies to identify optimal blocks from a mixed search space and determine the topology among heterogeneous blocks, ultimately constructing a novel architecture.
Additionally, a recent study, AutoCTS+~\cite{wu2023autocts+}, introduces a unified search strategy that incorporates both operators and hyperparameters, making it the first work to include hyperparameters in the search space.
However, the current automated approaches face challenges due to the high computational cost of neural architecture search, which limits their practical application and the exploration of a wider range of spatio-temporal operators at in finer granularity.

\section{Conclusion}
\label{sec:Conclusion}
In this paper, we propose AutoSTF, a novel decoupled neural architecture search framework for cost-effective automated spatio-temporal forecasting. 
By decoupling the mixed search space into temporal and spatial space, we design representation compression and parameter-sharing schemes to mitigate the parameter explosion in our framework.
The decoupled spatio-temporal search not only expedites the model optimization process but also leaves new room for more effective spatio-temporal dependency modeling. 
In addition, in the multi-patch transfer module, we aim to jointly capture multi-granularity temporal dependencies and extend the spatial search space to enable finer-grained layer-wise spatial dependency search.
In spatial search, we design layer-wise message-passing spatial operators to capture the spatio-temporal dependencies by automatically selecting the optimal adjacency matrix in different layers, thereby enhancing forecasting accuracy.
Comprehensive experiments conducted on eight datasets highlight the superior performance of AutoSTF in terms of both accuracy and efficiency.
In future work, focusing on improving the search algorithm, expanding the search space, and scaling the model to handle large datasets presents a valuable and meaningful challenge. 
Additionally, developing a method to automatically determine the optimal number of patches for different datasets is another meaningful task.

\bibliographystyle{ACM-Reference-Format}
\bibliography{sample-base}

\appendix
\clearpage
\newpage

\section{Model Design and Analysis}
\label{Appendix:sec:Model Design and Analysis}

In this section, we first introduce existing automated spatio-temporal forecasting models within a generic framework. 
Then, we examine the efficiency and effectiveness bottlenecks of existing approaches and elaborate on our model's design intuition.

\begin{figure}[h]
  \centering
  \includegraphics[width=\linewidth]{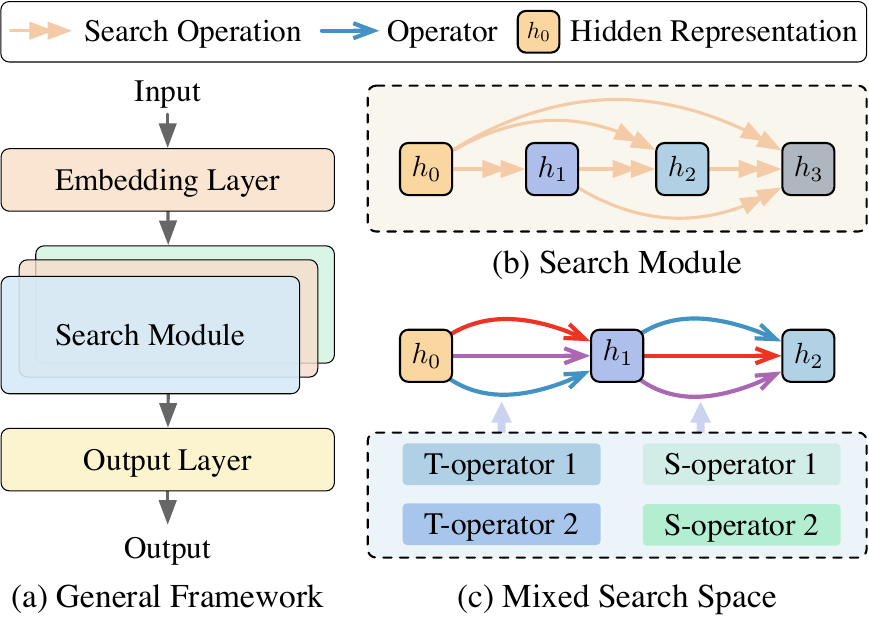}
  \caption{An overview of existing automated spatio-temporal forecasting models. (a) and (b) represent the general framework and search module of automated models, respectively. Meanwhile, (c) denotes the mixed search space, encompassing both temporal and spatial operators.}
  \label{fig:generic_framework}
  \Description{}
  \vspace{-0.2cm}
\end{figure}

\subsection{Automated Spatio-temporal Models}
\textbf{General Framework.}
In order to gain a deeper understanding of the potential of automated spatio-temporal forecasting models, we investigate several representative works ~\cite{wu2021autocts,pan2021autostg,li2022autost,jin2022automated}. 
We first provide an overview of the general automated model, as shown in Figure~\ref{fig:generic_framework}, which primarily comprises three key components, \ie the embedding layer, search module, and output layer.
Firstly, an embedding layer is responsible for converting raw time series data into latent representations or fusing various embeddings, such as positional, spatial, and temporal embeddings, as described by AutoST~\cite{li2022autost}. 
Secondly, the search module is the key component that may comprise several spatio-temporal blocks. Each block explores spatio-temporal dependencies by automatically examining the optimal combination of operators within a mixed search space.
Finally, an output layer integrates the latent representations derived from the preceding layer, resulting in forecasting.

\textbf{Mixed Search Space.}
We summarize the search space of spatio-temporal blocks in existing representative models~\cite{wu2021autocts,pan2021autostg,li2022autost,jin2022automated}.
By following previous studies~\cite{wu2021autocts,lai2023lightcts}, we classify frequently used operators for automated spatio-temporal forecasting and analyze the time and space complexity of each operator, as shown in Table~\ref{tab:search_space}.
Specifically, within the field of automated spatio-temporal forecasting, we categorize temporal operators into two groups. (1) The Convolutional Neural Network (CNN) family, such as the Temporal Convolutional Network (TCN), employs dilated causal convolutions to process spatio-temporal data. (2) The Transformer family, which employs self-attention mechanisms to facilitate the weights among input time steps, enhancing the representation ability of temporal information over extended sequences.
Similarly, we classify spatial operators into two families. (1) The Graph Neural Network (GNN) family, which leverages predefined or adaptive adjacency matrices learned from a data-driven method. (2) The Transformer family, which applies attention mechanisms across different time series to capture high-order spatial dependencies.
\begin{table}[b] \footnotesize
  \caption{The operators in mixed search space of existing automated models. $\mathcal{T}$ and $\mathcal{S}$ denotes the temporal and spatial operators, respectively. $\Omega(time)$ and $\Omega(space)$ denotes the time and space complexity, respectively.}
  \label{tab:search_space}
  \begin{tabular}{c|cc|cc}
    \toprule
    Type & Family & Models & $\Omega(time)$ & $\Omega(space)$ \\
    \midrule
    \multirow{2}*{$\mathcal{T}$} 
    & CNN-based & \cite{wu2021autocts,wu2023autocts+,pan2021autostg,ke2023autostg+} & $\mathcal{O}(D^{2}\cdot N \cdot P)$ & $\mathcal{O}(D^{2})$ \\

    & Transformer & \cite{wu2021autocts,wu2023autocts+,li2022autost} & $\mathcal{O}(D\cdot N\cdot P\cdot (P+D))$ & $\mathcal{O}(D^{2})$ \\
    
    \midrule
    \multirow{2}*{$\mathcal{S}$} 
    & GNN-based & \cite{wu2021autocts,pan2021autostg,ke2023autostg+,li2022autost} & $\mathcal{O}(D\cdot N\cdot P\cdot (P+D))$ & $\mathcal{O}(D^{2})$ \\

    & Transformer & \cite{wu2021autocts,wu2023autocts+} & $\mathcal{O}(D\cdot N\cdot P\cdot (P+D))$ & $\mathcal{O}(D^{2})$ \\
    \bottomrule
  \end{tabular}
\end{table}

\begin{table}[b] \small
  \caption{Analysis of running time and parameter size of temporal and spatial operators in automated spatio-temporal forecasting models.}
  \label{tab:decouple_space}
  \begin{tabular}{c|c|cc}
    \toprule
    Model & Metric & Temoral Operator & Spatial Operator \\
    \midrule
    \multirow{2}*{AutoSTG \cite{pan2021autostg}} 
    & Run Time & 17.7\% & 82.3\% \\
    & Parameters & 22.4\% & 77.6\% \\
    
    \midrule
    \multirow{2}*{AutoCTS \cite{wu2021autocts}} 
    & Run Time & 37.3\% & 63.7\% \\
    & Parameters & 26.3\% & 73.7\% \\
    \bottomrule
  \end{tabular}
\end{table}

\begin{figure}[t]
  \centering
  \includegraphics[width=1\linewidth]{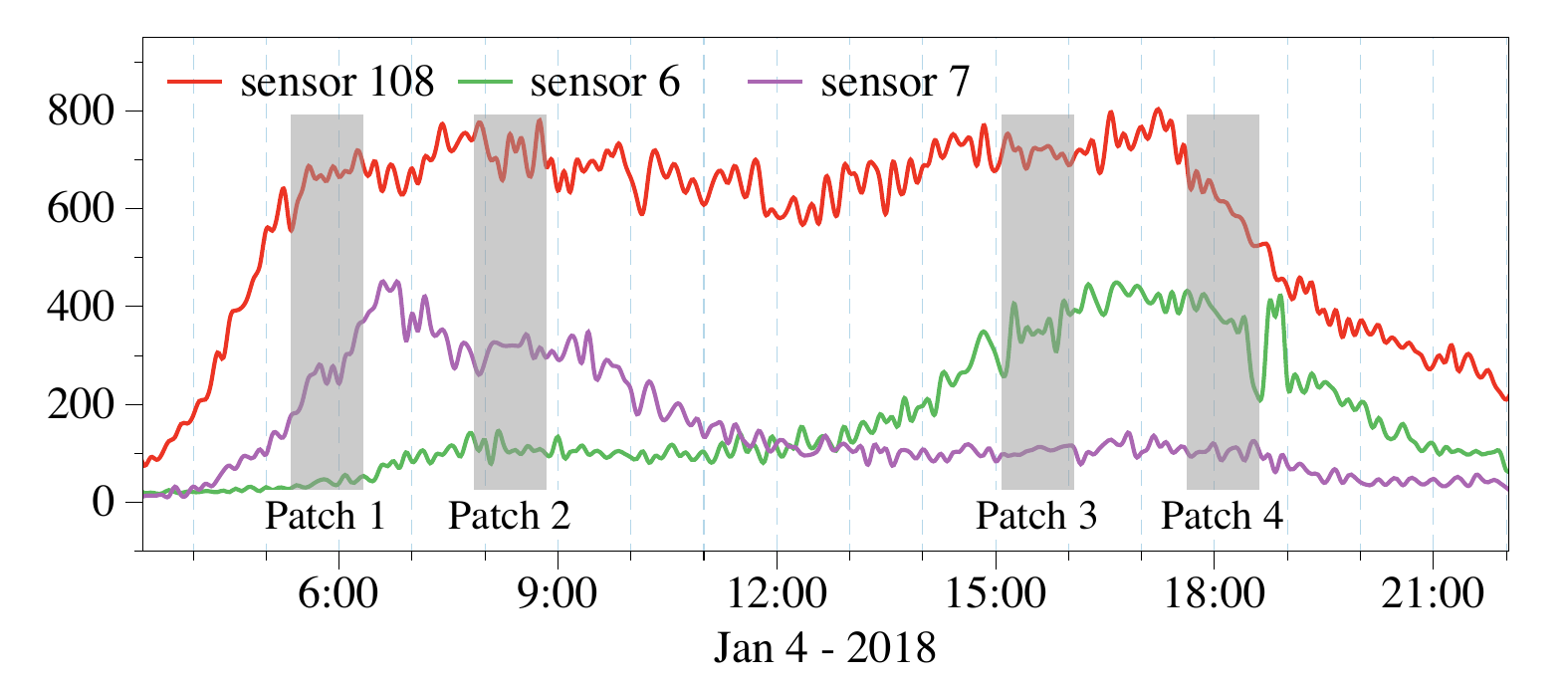}
  \vspace{-0.4cm}
  \caption{An example of the different traffic patterns on the PEMS04 dataset. Spatial dependencies may vary based on the time intervals of varying granularity.}
  \label{fig:multi-patch}
  \Description{}
  \vspace{-0.2cm}
\end{figure} 

\subsection{The Efficiency Bottleneck}
\label{sec:Decoupled the Search Space}

To investigate the efficiency bottleneck of temporal and spatial operators in the mixed search space, we analyze run time and parameters via a case study. 
Specifically, we select two state-of-the-art automated spatio-temporal forecasting models, namely AutoSTG~\cite{pan2021autostg} and AutoCTS \cite{wu2021autocts}. 
We conduct the experiment with the PEMS-BAY dataset and use the settings reported by the original papers, and the results are shown in Table \ref{tab:decouple_space}. 
We calculate the run time and parameters of each operator and then count the proportions of temporal and space operators in mixed search space. 
As reported, the spatial operator typically has a larger parameter size and consumes more computational resources than the temporal operator.
Such observations lead to two potential optimization directions: reducing computational redundancies and reducing the number of parameters involved in spatial operators.

Firstly, it is evident that temporal information is crucial for modeling dependencies within each time series but may be redundant when exploring dependencies across different locations~\cite{lai2023lightcts}. 
Thus, the first critical insight is that decoupling the mixed search space into temporal space and spatial space and devising a representation compression scheme to distill the key temporal information for spatial search can speed up the overall computations.
As presented in Table \ref{tab:search_space}, the time and space complexities of all temporal and spatial operators are positively correlated with the numbers of sensors $N$, historical time steps $P$, and hidden dimension $D$. 
However, $N$ is decided by the raw time series data and should be left as a constant in the spatio-temporal forecasting model. 
Directly reducing $D$ can inevitably degrade forecasting accuracy, as validated by previous studies~\cite{chen2020multi,wu2020connecting}. 
Fortunately, recent studies~\cite{lai2023lightcts} prove $P$ also can not be directly reduced during the temporal modeling phase but can be compressed in the spatial modeling phase without influencing the forecasting performance. 
Thus, it is intuitive to devise a representation compression scheme to accelerate the overall neural architecture search. 

Secondly, existing automated forecasting models involve a substantial number of parameters in the spatial search process (as shown in Table~\ref{tab:decouple_space}), inducing considerable computation overhead. To tackle this issue, we propose to share parameters among the same type of spatial blocks to reduce the spatial search overhead.

In this paper, we investigate decoupling the mixed spatio-temporal search space and respectively devise representation compression and parameter-sharing schemes to optimize the neural architecture search efficiency.

\begin{table}[t] \small
  \caption{
  The classification of models based on different types of adjacency matrices.
  }
  \label{tab:optimal_spatial_operation}
  \begin{tabular}{c|cc}
    \toprule
    Type & Matrix & Models \\
    \midrule
    \multirow{3}*{\rotatebox{0}{FF}} 
    &\makecell[c]{Distance-based \\ Matrix} & \makecell[c]{ DCRNN \cite{li2017diffusion}, GrapWaveNet \cite{wu2019graph}, \\ DGCN \cite{pan2019urban}, STGNN \cite{wang2020traffic}, \\ GMAN \cite{zheng2020gman}, ST-GRAT \cite{park2020st}, STDN \cite{yu2018spatio} }\\
    
    \cline{2-3}
    
    & Binary Matrix & \makecell[c]{ DGCN \cite{pan2019urban}, ASTGCN \cite{guo2019attention}, \\ STSGCN \cite{song2020spatial}, LSGCN \cite{huang2020lsgcn} }\\

    \cline{2-3}
    
    & DTW Matrix & STFGNN\cite{li2021spatial}, Auto-DSTSGN\cite{jin2022automated} \\

    \midrule
    \rotatebox{0}{SA}
    & Adaptive Matrix & \makecell[c]{ GrapWaveNet \cite{wu2019graph}, SLCNN \cite{zhang2020spatio}, \\ AGCRN \cite{bai2020adaptive}, MTGNN \cite{wu2020connecting}, STDN \cite{yu2018spatio} }\\
    
    \midrule
    \rotatebox{0}{FA}
    & Attention Matrix & \makecell[c]{ GeoMAN \cite{liang2018geoman}, LightCTS \cite{lai2023lightcts} }\\
    \bottomrule
  \end{tabular}
  \vspace{-0.3cm}
\end{table}

\subsection{The Effectiveness Bottleneck}
\label{sec:The Effectiveness Bottleneck}

From the effectiveness perspective, existing automated models explore spatio-temporal dependencies in a coarse-grained manner, leading to unsatisfied prediction accuracy.
As shown in Figure~\ref{fig:multi-patch}, the traffic patterns of three geo-distributed sensors are vastly different and exhibit diverged correlations. 
Specifically, Sensor 108 experiences high traffic flow throughout the day, Sensor 7 has the most significant traffic flow during morning peak hours, and Sensor 6 has the largest traffic flow during evening peak hours.
However, it has strong correlations in a finer granularity. 
For instance, the traffic flow in Patch 1 exhibits an increasing trend, while the traffic flow in Patch 4 shows a decreasing trend.
Thus, fine-grained exploration of spatial search can be beneficial in enhancing performance.

In addition, existing automated methods have employed the same message-passing scheme across different GNN layers, which fails to consider the unique spatial dependencies present in different multi-hop neighbors.
Recent advances have proposed diverse adjacency matrices for spatial dependency modeling and demonstrated the effectiveness of different adjacency matrices in different contexts ~\cite{jin2022automated}.
As summarized in Table \ref{tab:optimal_spatial_operation}, we categorize the commonly used adjacency matrices in three categories~\cite{li2023dynamic}:
(1) the fully fixed (\textbf{FF}) matrix that is the predefined adjacency matrix and fixed in both the training and inferencing phase;  
(2) the semi-adaptive (\textbf{SA}) matrix that the parameters of the adjacency matrix are learnable with the training phase, and fixed in the inferencing phase;
(3) the fully adaptive (\textbf{FA}) matrix that was learned from data-driven methods and adaptive in both the training and inferencing phase.
In fact, each adjacency matrix category can be regarded as a new class of spatial operators.
It is intuitive to incorporate varying adjacency matrix operators in different layers so as to encompass more spatial semantics for more effective forecasting.
In this work, we construct an extended spatial search space that includes the distance matrix, adaptive matrix, and attention matrix operators.
Based on the new spatial space, we propose a more flexible message-passing scheme to automatically select the most suitable adjacency matrix in different layers for better forecasting accuracy. 

\section{Supplementary of Experiment}
\label{appendix:Supplementary of Experiments}

\subsection{Datasets}
\label{appendix:Datasets}

We conducted experiments on eight benchmark datasets for both multi- and single-step forecasting. 
These datasets can be categorized as traffic speed, traffic flow, solar-energy, and electricity.

\textit{Multi-step forecasting.}
The METR-LA and PEMS-BAY datasets are both traffic speed datasets released by Li et al. \cite{li2017diffusion}. 
The PEMS03, PEMS04, PEMS07, and PEMS08 datasets are all traffic flow collected from the Caltrans Performance Measurement System (PeMS) and released by Song et al. \cite{song2020spatial}. 

\textit{Single-step forecasting.}
The solar-energy dataset captures the 10-minute variations of 137 PV plants across Alabama.
The electricity dataset reflects the hourly electricity consumption of 321 clients from 2012 to 2014.
These two datasets are pre-processed and released by Lai et al. \cite{lai2018modeling}.

\begin{table}[t] \small
  \caption{The operators of temporal search space.}
  \vspace{-0.1cm}
  \label{tab:temporal_operators}
  \begin{tabular}{c|c}
    \toprule
    $\mathcal{T}$ Operators & Formulary \\
    
    \midrule
    GDCC & $H^{(i)} = (Z^{(i)} * W_{1}) \odot \sigma(Z^{(i)} * W_{2})$ \\

    \midrule
    Informer & $H^{(i)} = Softmax(\frac{smp(Z^{i}W_{Q})(Z^{i}W_{K})^{T}}{\sqrt{D^{\prime}}})(Z^{i}W_{V})$ \\

    \bottomrule
  \end{tabular}
  \vspace{-0.4cm}
\end{table}

\begin{table}[t] \small
  \caption{Statistic of datasets.}
  \label{tab:dataset}
  \vspace{-0.2cm}
  \begin{tabular}{c|ccc|ccc}
    \toprule
    Dataset & Nodes & Time steps & Interval & Split & P & Q \\
    \midrule
    METR-LA & 207 & 34,272 & 5 min & 7:1:2 & 12 & 12 \\
    PEMS-BAY & 325 & 52,116 & 5 min & 7:1:2 & 12 & 12 \\
    PEMS03 & 358 & 26,208 & 5 min & 6:2:2 & 12 & 12 \\
    PEMS04 & 307 & 16,992 & 5 min & 6:2:2 & 12 & 12 \\
    PEMS07 & 883 & 28,224 & 5 min & 6:2:2 & 12 & 12 \\
    PEMS08 & 170 & 17,856 & 5 min & 6:2:2 & 12 & 12 \\
    \midrule
    Solar-energy & 137 & 52,560 & 10 min & 6:2:2 & 168 & 1 \\
    Electricity & 321 & 26,304 & 1 hour & 6:2:2 & 168 & 1 \\

    \bottomrule
  \end{tabular}
  \vspace{-0.4cm}
\end{table}

\subsection{Baselines}
\label{appendix:Baselines}
In this section, we provide a brief overview of the baseline methods for reference in our paper. 

\begin{itemize}
\item DCRNN \cite{li2017diffusion}: 
A diffusion convolutional recurrent neural network for traffic forecasting that incorporates spatio-temporal dependency based on distance adjacency matrix.

\item STGCN \cite{yu2018spatio}: 
A spatio-temporal graph convolutional network based on the distance adjacency matrix for traffic forecasting, which captures spatio-temporal correlations through modeling multi-scale traffic networks.

\item Graph WaveNet \cite{wu2019graph}: 
It uses adaptive and predefined adjacency matrices to explore the spatio-temporal dependency.

\item AGCRN \cite{bai2020adaptive}: 
An adaptive graph convolution recurrent network with adaptive matrix to make forecasting.

\item MTGNN \cite{wu2020connecting}: 
The model is a general graph neural network that is specifically designed for analyzing multivariate time series data. It is based on an adaptive matrix approach.

\item STID \cite{shao2022spatial}: 
A simple yet effective baseline for Multivariate Time Series forecasting by incorporating spatial and temporal identity information.

\item STHODE \cite{yao2023spatio}: 
It employs a spatial-temporal hypergraph coupled with ODE networks to enhance traffic prediction by capturing high-order dependencies in road networks and traffic dynamics.

\item DeepSTUQ \cite{qian2023uncertainty}: 
This model employs a spatio-temporal model for traffic data's complex correlations and uses two sub-networks for aleatoric uncertainty, while integrating Monte Carlo dropout and Adaptive Weight Averaging for epistemic uncertainty.

\item STJGCN \cite{zheng2023spatio}: 
It offers accurate traffic forecasting by constructing dynamic spatio-temporal joint graphs and applying dilated graph convolutions with a multi-range attention mechanism to capture dependencies across multiple time steps and ranges.

\item STAEformer \cite{liu2023spatio}: 
A novel component, termed as the Spatio-Temporal Adaptive Embedding, is introduced, demonstrating its capability to yield outstanding results when integrated with vanilla transformers.

\item LightCTS \cite{lai2023lightcts}: 
A lightweight approach that utilizes a plain stacking schema to explore spatio-temporal correlations. The model is based on an attention matrix, which allows for the identification of relevant features in the data.

\item AutoSTG \cite{pan2021autostg}: 
An automated spatio-temporal graph forecasting method that utilizes meta-learning to train an adaptive matrix for Graph Convolutional Networks (GCN).

\item AutoCTS+ \cite{wu2023autocts+}:
A joint and scalable approach that designs a search space containing both operators and hyperparameters to devise effective time series forecasting automatically.

\item AutoCTS \cite{wu2021autocts}:
This automated forecasting model is designed with both a micro and macro search space to model possible architectures for spatio-temporal dependencies.
\end{itemize}

\subsection{Evaluation Metrics.}
\label{appendix:Evaluation Metrics}
We followed previous works \cite{wu2021autocts,pan2021autostg,wu2023autocts+} and adopted different metrics to evaluate the model's performance in multi- and single-step forecasting. 
Specifically, Mean Absolute Error (MAE), Root Mean Squared Error (RMSE), and Mean Absolute Percentage Error (MAPE) were utilized to assess the accuracy of multi-step forecasting. 
For single-step forecasting, we utilized Root Relative Squared Error (RRSE) and Empirical Correlation Coefficient (CORR) as evaluation metrics. 
Lower values of MAE, RMSE, MAPE, and RRSE are indicative of higher accuracy, while larger CORR values denote higher accuracy.

\subsection{Implementation Details}
\label{appendix:Implementation Details}

The experiments were conducted on three different server configurations. 
The first configuration consisted of three Linux Centos servers, each equipped with 4 RTX 3090 GPUs. 
The second configuration included one Linux Ubuntu server with 2 V100 GPUs, and the third configuration included one Linux Ubuntu server with 2 A800 GPUs.

Following the previous works \cite{wu2021autocts,pan2021autostg,liu2018darts}, we use MAE as the loss function to train the model and use Adam with a learning rate of 0.001 for both the architecture parameters $\Theta$ and the network weights $\omega$. 
In addition, we use a weight decay of 0.0001 as the optimizer, and the batch size is set to 64. 
In the temporal and spatial DAG, we set the number of nodes as 4. 
In order to ensure reproducibility,  we have provided the detailed settings for each dataset in our source code.

\begin{figure}[t]
  \centering
  \includegraphics[width=1\linewidth]{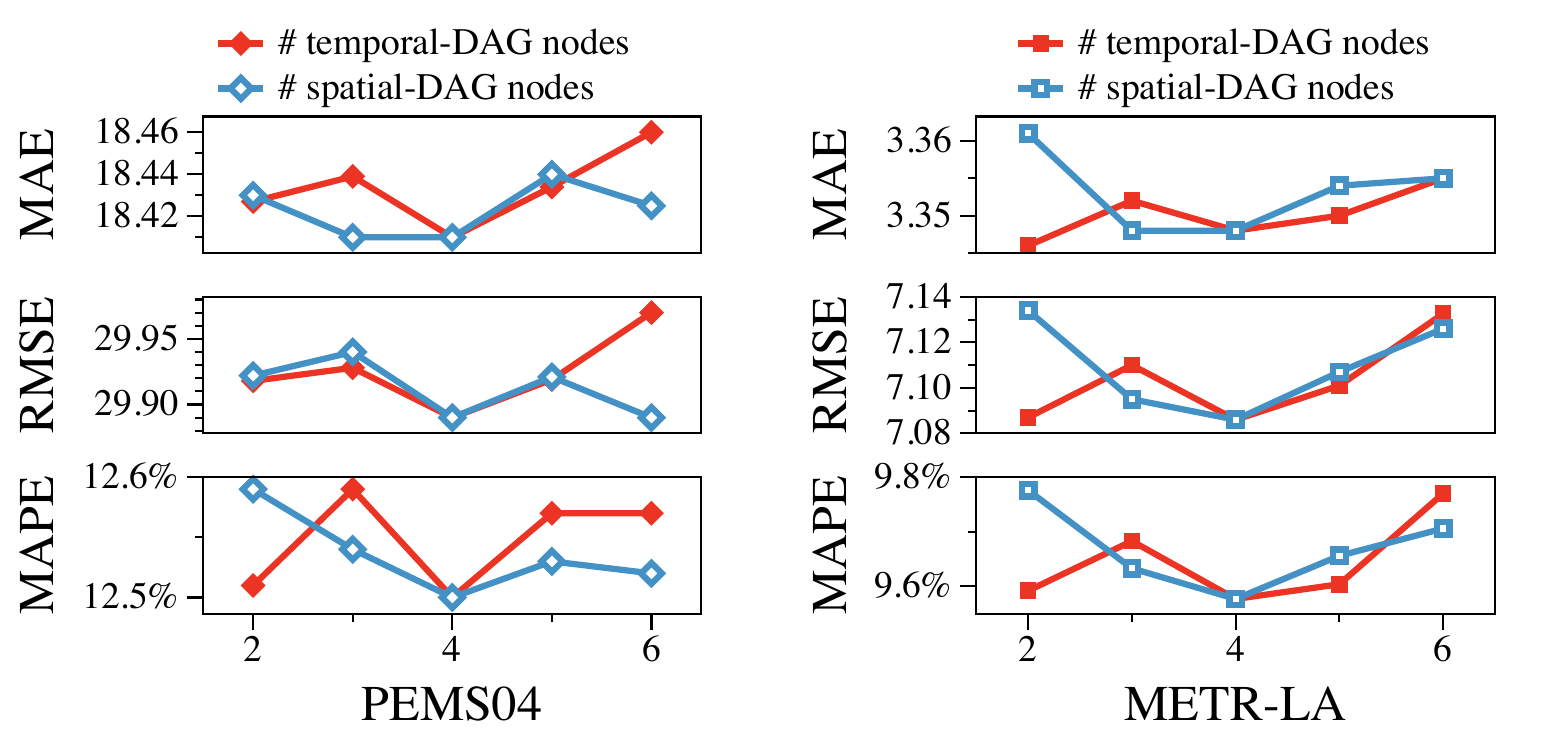}
  \caption{Parameter sensitivity analysis on PEMS04 and METR-LA datasets.}
  \label{fig:parameter_sensitivity}
  \Description{}
  \vspace{-0.4cm}
\end{figure}

\subsection{Parameter Sensitivity Analysis.}
\label{appdix:Parameter Sensitivity Analysis}
In this experiment, we evaluate the impact of key hyperparameters in AutoSTF, including $N_{\mathcal{T}}$, which represents the number of nodes in the temporal-DAG, $N_{\mathcal{S}}$, which represents the number of nodes in the spatial-DAG, and $M$, which represents the number of patches in the multi-patch transfer module.
We set the default values for $N_{\mathcal{T}}$, $N_{\mathcal{S}}$, and $M$ to 4, 4, and 2, respectively. 
For instance, when $M=2$, we divide the temporal embedding $H_{\mathcal{T}}$ into two patches.
We vary $N_{\mathcal{T}}$ and $N_{\mathcal{S}}$ among \{2, 3, 4, 5, 6\}, and $M$ among \{1, 2, 3, 4, 6\}, while keeping the remaining hyperparameters at default values.
We present the results in Figures \ref{fig:parameter_sensitivity} and Table \ref{tab:scale_sensitivity}, where the utilized datasets represent different types of data. Specifically, METR-LA corresponds to traffic speed data, while PEMS04 corresponds to traffic flow data.

As shown in Figure \ref{fig:parameter_sensitivity}, AutoSTF achieves the best accuracy under $N_{\mathcal{T}}$ = 4 and $N_{\mathcal{S}}$ = 4. 
Reducing $N_{\mathcal{T}}$ or $N_{\mathcal{S}}$ may result in insufficient expressiveness of the model, leading to a reduction in the accuracy of the model. 
Increasing $N_{\mathcal{T}}$ or $N_{\mathcal{S}}$ augments the complexity of the temporal and spatial search space, which may lead to overfitting problems.
As a result, it may slightly degrade the accuracy. 
Additionally, we present the results for different numbers of patches in Table \ref{tab:scale_sensitivity}. It can be observed that the impact of patches with varying granularities varies significantly across different datasets. 
For instance, in traffic flow datasets, the fluctuations in traffic flow at different time points might be more pronounced. 
Therefore, examining spatio-temporal correlations with a more fine-grained approach can enhance the accuracy of the model. 
For the PEMS04 dataset, using a larger value of $M$ can lead to an improvement in the model's accuracy. 
On the other hand, in traffic speed datasets, the average speed of vehicles tends to change less over a short period of time. 
Thus, employing fewer patches to explore spatio-temporal dependencies may yield more accurate results. 
For the METR-LA dataset, a smaller value of $M$ achieves the best performance compared to larger values of $M$.

\begin{table}[t]\small
  \caption{The analysis of patch numbers on PEMS04 and METR-LA datasets.}
  \label{tab:scale_sensitivity}
  \begin{tabular}{c|c|ccc|c}
    \toprule

    \text { Dataset } & \text { \# patch }  & \text { MAE } & \text { RMSE } & \text { MAPE } & \text { Parameters}\\

    \midrule
    \multirow{5}{*}{\rotatebox{90}{METR-LA}}
    & \textbf{ 1 }     & \textbf{3.35} & \textbf{7.09} & \textbf{9.58}   & 381,315 \\
    & \text{ 2 }    & 3.37 & 7.13 & 9.80   & 382,315 \\
    & \text{ 3 }      & 3.39 & 7.15 & 9.72  & 383,337 \\
    & \text{ 4 }      & 3.38 & 7.18 & 9.82  & 384,360 \\
    & \text{ 6 }     & 3.40 & 7.24 & 9.88   & 386,407 \\

    \midrule
    \multirow{5}{*}{\rotatebox{90}{PEMS04}}
    & \text{ 1 }      & 18.67 & 30.27 & 12.87 & 514,257 \\
    & \text{ 2 }    & 18.44 & 29.93 & 12.76 & 515,275 \\
    & \textbf{ 3 }      & \textbf{18.38} & \textbf{29.86} & \textbf{12.58} & 516,297 \\
    & \text{ 4 }      & 18.44 & 29.91 & 12.61 & 517,320 \\
    & \text{ 6 }     & 18.41 & 29.92 & 12.56 & 519,367 \\
    \bottomrule
  \end{tabular}
  \vspace{-0.2cm}
\end{table}

\subsection{Case Study}
\label{appdix:Case Study}
\begin{figure}[t]
  \centering
  \includegraphics[width=\linewidth]{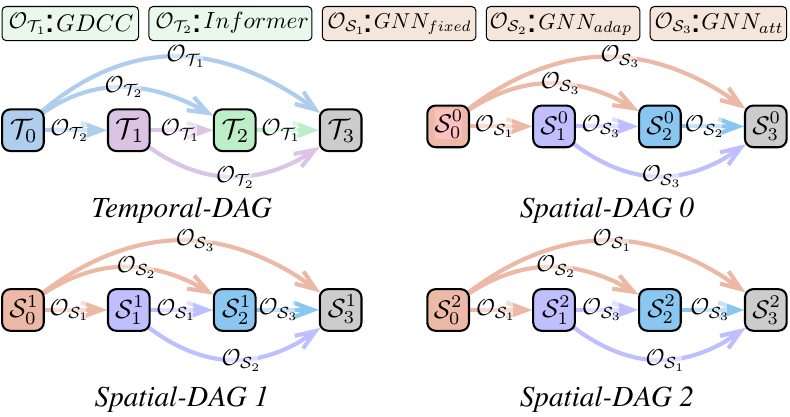}
  \caption{The visualization of the model architecture on the PEMS04 datasets. $\mathcal{T}_i$ and $\mathcal{S}_i^*$ are the hidden representations.}
  \label{fig:DAG_visualization}
  \Description{}
\end{figure}

Due to space limitations, we only show the neural architecture predicted by AutoSTF for the PEMS04 dataset in Figure \ref{fig:DAG_visualization}. 
From the results, we can observe that in the temporal-DAG, the model automatically selects three $GDCC$ and three $Informer$ modules to explore the temporal dependencies. 
Furthermore, in the spatial search, our model has three spatial-DAGs. Based on the results, we can conclude that our model is capable of automatically designing appropriate message-passing aggregations for different spatial-DAGs in order to examine spatio-temporal dependencies.
This demonstrates the effectiveness of decoupling the mixed search space and performing fine-grained spatial search for different patches. Ultimately, this approach results in the development of models that excel in both efficiency and accuracy.

\begin{figure}[t]
  \centering
  \includegraphics[width=\linewidth]{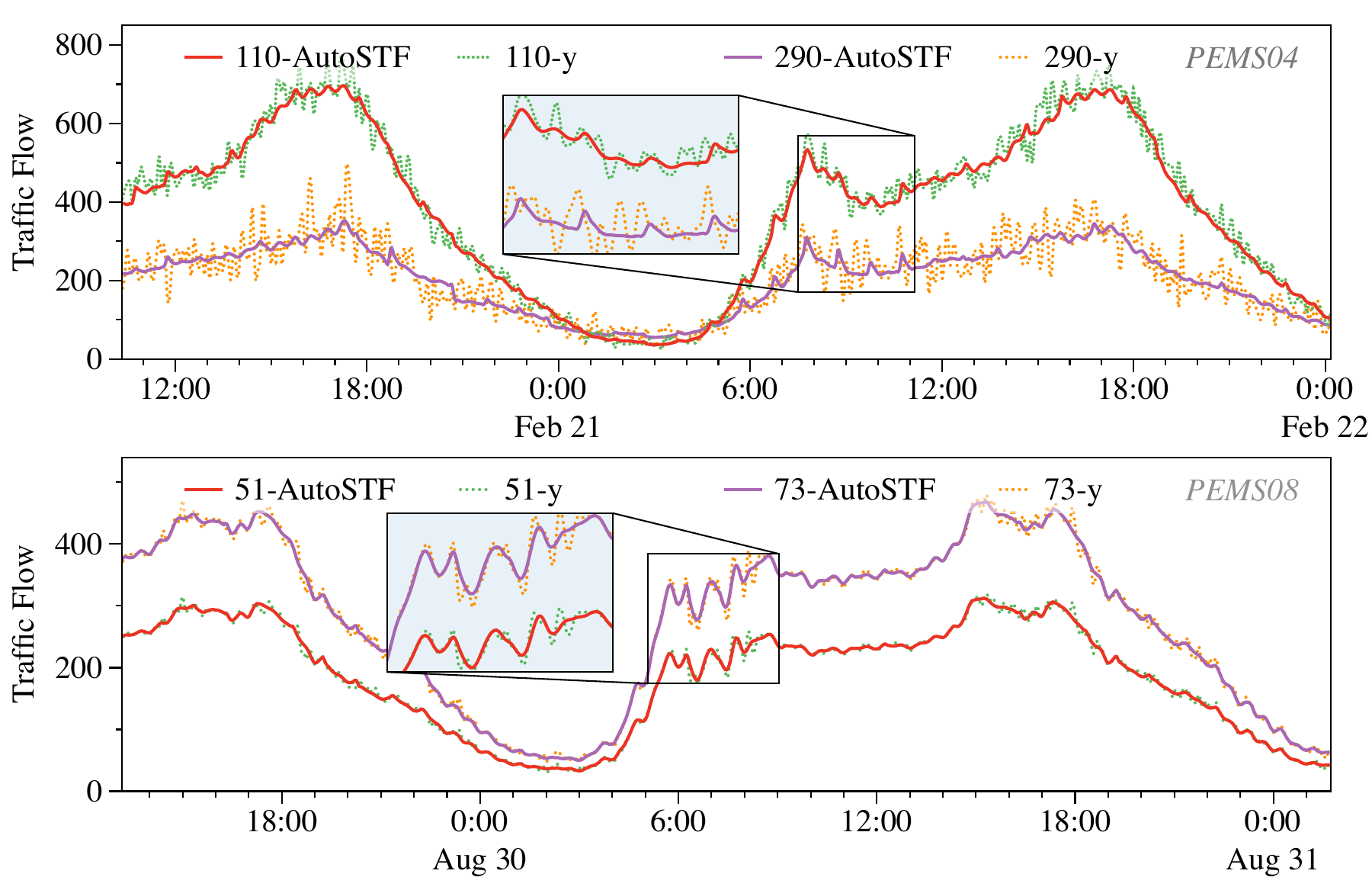}
  \caption{
  The visualization of the forecasting performance on the PEMS04 and PEMS08 datasets.}
  \label{fig:spatio-temporal case}
  \Description{}
\end{figure}

Additionally, we also conduct a case study to demonstrate our model's ability to capture and model spatio-temporal dependencies accurately, shown in Figure \ref{fig:spatio-temporal case}. 
We randomly select two sensors that are interconnected in PEMS04 and PEMS08 dataset, respectively.
We can observe from Figure \ref{fig:spatio-temporal case} that each two sensor shows a significant correlation, indicating that the traffic flow of one sensor correlates with the other sensor. 
For example, Sensor 290 in PEMS04 shows a strong spatio-temporal correlation with Sensor 110, when traffic flow is increased at Sensor 110, Sensor 290 will also increase. They exhibit a similar pattern, indicating a strong spatio-temporal correlation. 
Based on Figure \ref{fig:spatio-temporal case}, it is evident that our AutoSTF model accurately predicts the traffic flow of Sensor (110, 290) and (51, 73), showcasing remarkable accuracy in forecasting similar patterns. This demonstrates that AutoSTF can also effectively preserve the spatio-temporal correlation.

\subsection{The Improvement of AutoSTF}
\label{appdix:The Improvement of AutoSTF}

\begin{figure}[t]
  \centering
  \includegraphics[width=1\linewidth]{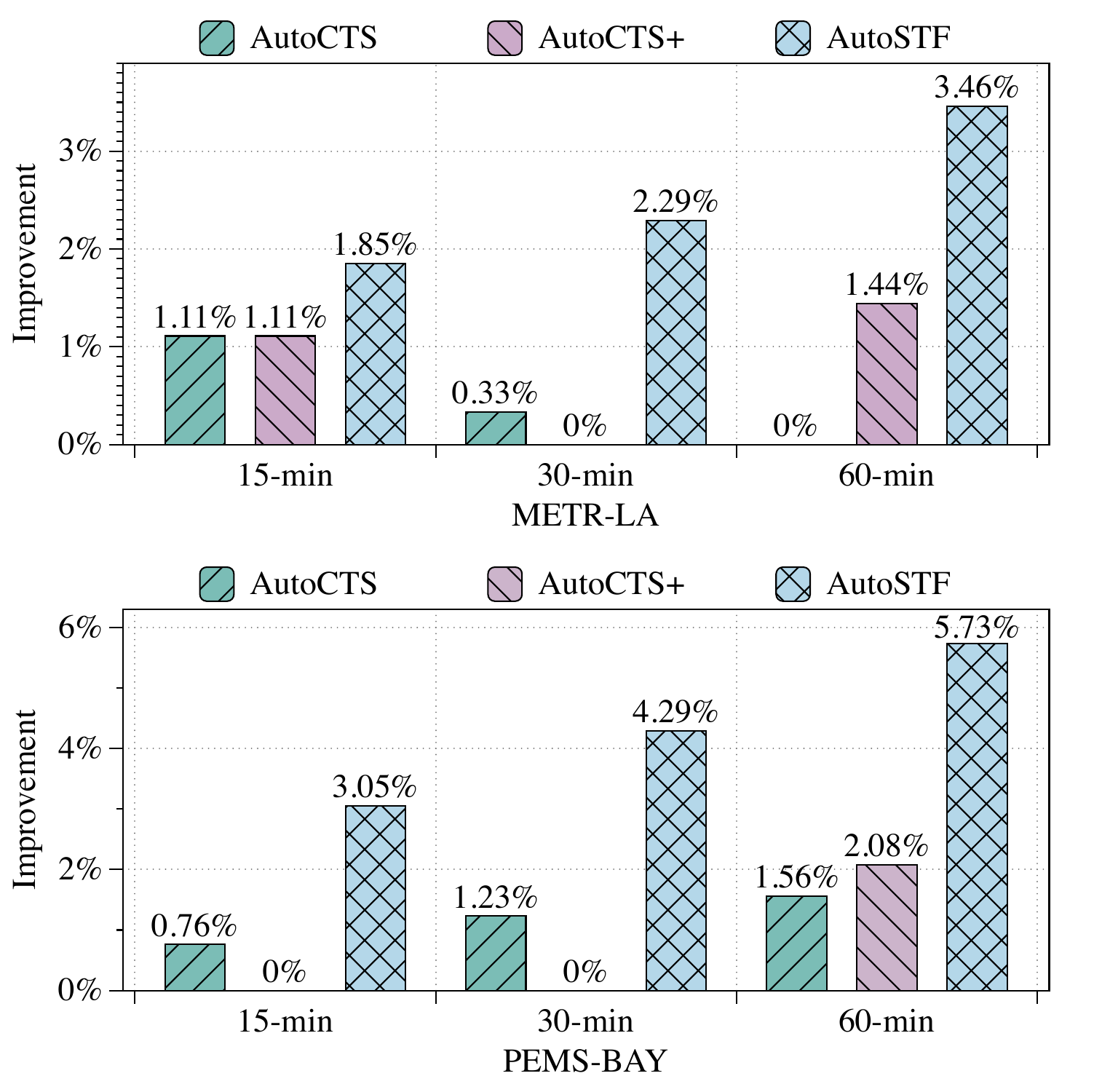}
  \vspace{-0.6cm}
  \caption{
  The visualization of the percentage improvement in MAE achieved by AutoSTF, AutoCTS+, and AutoCTS over the AutoSTG on MAE of METR-LA and PEMS-BAY datasets.}
  \label{fig:improve-AutoSTG}
  \Description{}
  \vspace{-0.4cm}
\end{figure}

To more clearly compare the performance improvements of AutoSTF and other NAS-based spatio-temporal forecasting models, we have calculated their performance improvements relative to AutoSTG based on the MAE on METR-LA and PEMS-BAY datasets. 
%
As observed in Figure \ref{fig:improve-AutoSTG}, it is evident that AutoSTF exhibits significant improvements in all different time-scale predictions compared to other automated spatio-temporal forecasting models.

\subsection{Complexity Analysis}
\label{appdix:Complexity Analysis}
In this section, we analyze the algorithmic complexity of AutoSTF. 
In AutoSTF, we compute the maximum space and time complexity that occurs during the neural architecture search process. Assuming AutoSTF comprises one temporal-DAG and 
$m$ spatial-DAGs, each containing $k$ edges (search operation), where $m$ is the number of patches.
In the temporal search module, we hypothesize that each search operation selects an $informer$ (the most time-consuming operator in our search space), with the space complexity of the $informer$ being $O(TlogT)$ and the time complexity being $O(TlogT)$ , where $T$ is the sequence length. Thus, the total space complexity during the temporal search phase is $O(kTlogT)$ , and the total time complexity is $O(kTlogT)$ .
In the spatial search module, the most time-consuming operator is $GNN_{att}$, which is based on the spatial transformer architecture. It has a space complexity of $O(N^2)$ and a time complexity of $O(N^2)$, where $N$ is the number of nodes. Consequently, the overall space complexity in the spatial search phase amounts to $O(mkN^2)$, and the overall time complexity reaches $O(mkN^2)$.
We can determine that the space and time complexities of the neural architecture search of AutoSTF are respectively $O(kTlogT+mkN^2)$ and $O(kTlogT+mkN^2)$.

\begin{table}[t] \small
  \caption{The comparison of forecasting accuracy and search efficiency between AutoSTF in decoupled search space and AutoSTF-Mixed in mixed search space.}
  \label{appendix:traffic_flow}
  \begin{tabular}{c|c|cc}
    \toprule

    Datasets & Metric & AutoSTF-Mixed &  AutoSTF \\
    
    \midrule
    \multirow{5}{*}{\text { PEMS03 }}
    & \text{ MAE }  & 15.05   & \textbf{14.44}\\
    & \text{ RMSE } & 24.78   & \textbf{23.94}\\
    & \text{ MAPE } & 15.09\% & \textbf{13.79}\%\\
    \cline{2-4}
    & \makecell[c]{Search Time \\ (s/epoch)} & 187 & \textbf{109}\\
    
    \midrule
    \multirow{5}{*}{\text { PEMS04 }}
    & \text{ MAE }  & 19.12   & \textbf{18.38}\\
    & \text{ RMSE } & 30.61   & \textbf{29.86}\\
    & \text{ MAPE } & 13.20\% & \textbf{12.58\%}\\
    \cline{2-4}
    & \makecell[c]{Search Time \\ (s/epoch)} & 173 & \textbf{112}\\

    \midrule
    \multirow{5}{*}{\text { PEMS07 }}
    & \text{ MAE }  & 19.86   & \textbf{19.50}\\
    & \text{ RMSE } & 32.92   & \textbf{32.66}\\
    & \text{ MAPE } & 8.33\%  & \textbf{8.14\%}\\
    \cline{2-4}
    & \makecell[c]{Search Time \\ (s/epoch)} & 623 & \textbf{465}\\

    \midrule
    \multirow{5}{*}{\text { PEMS08 }}
    & \text{ MAE }  & 14.77   & \textbf{14.07}\\
    & \text{ RMSE } & 23.69   & \textbf{23.17}\\
    & \text{ MAPE } & 9.46\%  & \textbf{9.14\%}\\
    \cline{2-4}
    & \makecell[c]{Search Time \\ (s/epoch)} & 144 & \textbf{107}\\
    
    \bottomrule
  \end{tabular}
  \vspace{-0.2cm}
\end{table}

\section{The Analysis of Decoupled Search Space}
\label{appdix:The Analysis of Decoupled Search Space}

In this section, we conducted a comprehensive analysis and experiments of decoupled the search space, which involved comparing the model's performance and efficiency before and after decoupling (Appendix \ref{append:The accuracy and efficiency of AutoSTF compared with AutoSTF-mixed}), as well as its ability to maintain spatiotemporal correlations (Appenddix \ref{appendix:The analysis of the spatio-temporal modeling ability of AutoSTF}).

\subsection{The accuracy and efficiency of AutoSTF compared with AutoSTF-mixed}
\label{append:The accuracy and efficiency of AutoSTF compared with AutoSTF-mixed}

We conducted experiments to validate the decoupled search space, as shown in Table \ref{appendix:traffic_speed} and Table \ref{appendix:traffic_flow}. In our model variant, AutoSTF-mixed, we created a mixed search space that combines both temporal and spatial operators. We utilized two directed acyclic graphs (DAGs) to simulate the forward flow during the search and training phases of a neural network. 
Within these two DAGs, we searched for the optimal neural architecture for the AutoSTF-mixed model. We evaluated the forecasting accuracy and average search time per epoch to assess the performance of the AutoSTF-mixed model. As presented in Table \ref{appendix:traffic_speed} and Table \ref{appendix:traffic_flow}, it is surprising to observe that AutoSTF-mixed did not achieve the highest forecasting accuracy and efficiency compared to AutoSTF. 
This outcome can be attributed to the presence of numerous operators in the mixed search space. The weights assigned to operators by the DARTS algorithm may deviate significantly from their actual operator, resulting in a substantial performance gap (see AutoCTS). Conversely, searching within both temporal and spatial search spaces allows for more precise identification of the optimal operator, thereby enhancing the model's performance.

\begin{table*}[h]
  \caption{The comparison of forecasting accuracy and search efficiency between AutoSTF in decoupled search space and AutoSTF-Mixed in mixed search space on traffic speed datasets.}
  \label{appendix:traffic_speed}
  \vspace{-0.4cm}
  \begin{tabular}{c|c|ccc|ccc|ccc|c}
    \toprule

    \multirow{2}{*}{\text { Datasets }} & \multirow{2}{*}{\text { Models }} & \multicolumn{3}{c|}{15 ~min} & \multicolumn{3}{c|}{30 ~min} & \multicolumn{3}{c|}{60 ~min} & \multirow{2}{*}{\makecell[c]{ Search Time \\ (s/epoch) }}\\

    & & \text { MAE } & \text { RMSE } & \text { MAPE } & \text { MAE } & \text { RMSE } & \text { MAPE } & \text { MAE } & \text { RMSE } & \text { MAPE } \\
    
    \midrule

    \multirow{2}{*}{\text { METR-LA }}
    & \text{ AutoSTF-Mixed }  & 2.81 & 5.51 & 7.60\% & 3.16 & 6.49 & 9.12\% & 3.52 & 7.42 & 10.65\% & 247 \\
    & \text{ AutoSTF }     & \textbf{2.65} & \textbf{5.10} & \textbf{6.80\%} & \textbf{2.99} & \textbf{6.10} & \textbf{8.10\%} & \textbf{3.35} & \textbf{7.09} & \textbf{9.58\%} & \textbf{188} \\
    
    \midrule
    \multirow{2}{*}{\text { PEMS-BAY }}
    & \text{ AutoSTF-Mixed }  & 1.30 & 2.78 & 2.73\% & 1.62 & 3.72 & 3.64\% & 1.89 & 4.41 & 4.44\% & 414 \\
    & \text{ AutoSTF }     & \textbf{1.27} & \textbf{2.70} & \textbf{2.64\%} & \textbf{1.56} & \textbf{3.59} & \textbf{3.48\%} & \textbf{1.81} & \textbf{4.25} & \textbf{4.23\%} & \textbf{233}\\

    \bottomrule
  \end{tabular}
\end{table*}

\begin{figure*}[t]
  \centering
  \includegraphics[width=\linewidth]{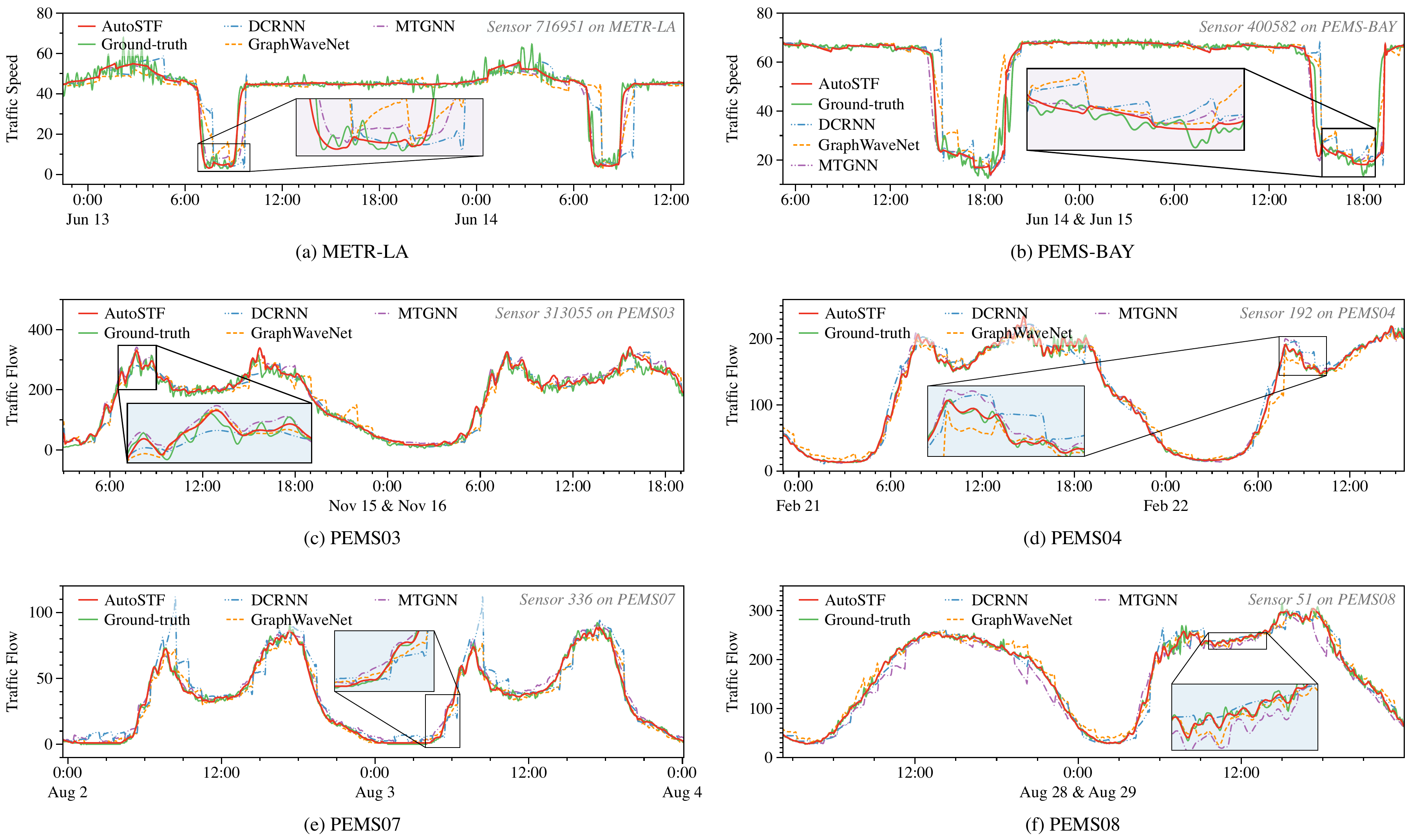}
  \vspace{-0.4cm}
  \caption{Visualization of AutoSTF and other classic manual-design models (DCRNN, GraphWaveNet, and MTGNN). The visualization illustrates the superior forecasting performance of AutoSTF compared to traditional models that utilize spatiotemporal correlation for improved prediction. This also demonstrates the capability of AutoSTF to preserve spatiotemporal correlation.}
  \label{appendix:fig:pred_vis}
  \Description{}
\end{figure*}

\subsection{The analysis of the spatio-temporal modeling ability of AutoSTF}
\label{appendix:The analysis of the spatio-temporal modeling ability of AutoSTF}

In this section, we provide a detailed analysis to demonstrate that even after decoupling spatio-temporal search into separate spatial and temporal searches, our model can still preserve spatio-temporal correlations.

Our model is developed as an end-to-end framework. After conducting the temporal search, the subsequent spatial search identifies the suitable operator based on the results of the temporal search. 
This selection process is optimized through backpropagation, ensuring the entire model framework is optimized holistically, thereby preserving the spatio-temporal correlation.
In addition, we design a more comprehensive spatial operator in spatial search space, including the fully fixed matrix, semi-adaptive matrix, and fully adaptive matrix. These operators can be adaptively chosen at different layers of the GNN, which can extract various complex spatiotemporal features.

To comprehensively evaluate the ability of AutoSTF to preserve spatio-temporal correlation, we compared AutoSTF with traditional methods such as MTGNN, GraphWaveNet, and DCRNN, which also claim to improve prediction performance by capturing spatio-temporal correlation.
For visualization purposes, we randomly selected a sensor from different datasets and displayed the results in Figure \ref{appendix:fig:pred_vis}. 
It is evident from Figure \ref{appendix:fig:pred_vis} that AutoSTF outperforms these classic spatio-temporal forecasting methods. 
This also demonstrates, from various perspectives, that AutoSTF has the ability to preserve spatio-temporal correlation, leading to enhanced prediction performance.
%

Overall, based on the detailed analysis and experiments presented above, our model is capable of preserving spatiotemporal correlations even after decoupling the spatiotemporal search into separate spatial and temporal searches.

\end{document}